%% file: main.tex
\documentclass[letterpaper,journal]{IEEEtran}
\usepackage{amsmath,amsfonts}
\usepackage{array}
\usepackage{textcomp}
\usepackage{stfloats}
\usepackage{url}
\usepackage{verbatim}
\usepackage{graphicx}
\usepackage{cite}
\usepackage{subcaption}
\usepackage[font=footnotesize]{caption}
\usepackage[table]{xcolor}
\usepackage{amsthm}
\usepackage{breqn}
\usepackage{tcolorbox}
\tcbuselibrary{breakable}
\usepackage{booktabs}
\usepackage{multirow}
\usepackage{tikz}
\usepackage{balance}
\usetikzlibrary{positioning, arrows.meta, calc, fit, backgrounds, shapes.geometric}

\newenvironment{note_p}{%
  \renewcommand{\abstractname}{Note to Practitioners}
  \begin{abstract}%
}{%
  \end{abstract}%
}

\newtcolorbox{contribbox}{breakable, colback=gray!8, colframe=black!70, boxrule=0.6pt, arc=1.5pt, left=4pt, right=4pt, top=3pt, bottom=3pt, fontupper=\small\itshape, before upper={\textbf{Our improvement:}\enspace}}
\newtcolorbox{summarybox}{breakable, colback=gray!8, colframe=black!70, boxrule=0.6pt, arc=1.5pt, left=4pt, right=4pt, top=3pt, bottom=3pt, fontupper=\small, before upper={\textbf{Summary:}\enspace}}

\begin{document}

\title{Occlusion-Aware, Quasi-Static, Stability-Oriented Trajectory Planning on Uneven Terrain}

\author{Amith Manoharan$^{1}$, Chinmay Mundane$^{1}$, Aayush Bahukhandi$^{2}$, K. Madhava Krishna$^{2}$,\\ Karel Zimmermann$^{3}$, and Arun Kumar Singh$^{1}$
\thanks{$^1$ are with the Institute of Technology, University of Tartu, Estonia.}
\thanks{$^2$ are with RRC, IIIT Hyderabad, India.}
\thanks{$^3$ is with the Czech Technical University, Prague.
\\contact email: amith.manoharan@ut.ee}
}




\maketitle

\begin{abstract}
Autonomous navigation in unstructured off-road environments requires reasoning about both vehicle--terrain interaction and environmental unknowns. We propose a model-based framework for generating quasi-static, stability-oriented reference trajectories for rigid, non-articulated four-wheeled vehicles on highly uneven terrain. Our work makes three primary contributions. First, we model blind spots caused by terrain occlusion as coverage-induced epistemic uncertainty in a fixed-feature Fourier terrain representation, quantified through a regularized inverse-Hessian estimate. Second, we propagate this uncertainty through the Nonlinear Least-Squares (NLS) pose/contact model using implicit differentiation and incorporate the resulting pose, contact-point, and per-wheel surface-normal uncertainty terms into trajectory optimization based on the Cross-Entropy Method (CEM). Third, we introduce a Flow Matching model that warm-starts terrain fitting, and we evaluate its fitting-accuracy--latency trade-off while retaining model-based refinement. Across six synthetic terrains with 30 matched start--goal pairs per terrain, the complete framework produced an observed failure rate of 18.9\%, compared with 46.1\% and 41.7\% for two representative baselines and 34.4\% for an ablation that removed the propagated-uncertainty scoring. Hardware evaluations span six distinct outdoor environments, with two representative executions presented in the paper and four additional executions included in the supplementary video. The evaluation also reports the accuracy--latency trade-off for the Flow Matching warm start.
\end{abstract}

\begin{note_p}
Autonomous field vehicles in agriculture, forestry, mining, and search and rescue often perceive only a fraction of the surrounding off-road terrain due to sensor occlusion. The planning pipeline must then decide whether to completely avoid unobserved areas or strike a balanced trade-off between visibility and stability. The former strategy can be overly conservative in real-world scenarios, particularly when all visible regions feature steep slopes that render traversal unsafe. Conversely, the balanced approach requires explicitly mapping missing terrain observations into uncertainty in vehicle pose and, consequently, trajectory planning cost. This paper directly tackles this problem, presenting a deployment-ready framework that runs in near real time on commodity onboard GPUs using only a limited-field-of-view depth camera. Evaluated in simulation and outdoor field trials on commercial wheeled platforms, the approach substantially reduces navigation failures compared to standard baselines without requiring expert demonstration datasets; the learned component serves strictly to accelerate map fitting while retaining explicit physical safety checks. Practitioners should note that the current formulation assumes a rigid four-wheeled platform operating at quasi-static speeds over 2.5D elevation surfaces, treating occlusion uncertainty as a tunable risk penalty rather than a hard constraint. The pipeline is readily deployable as a local rough-terrain path generator, with natural extensions to dynamic slip compensation, active sensor pointing, and adaptation to heavy tracked or articulated equipment.
\end{note_p}


\section{Introduction}

\begin{figure}
    \centering
    \includegraphics[width=\linewidth]{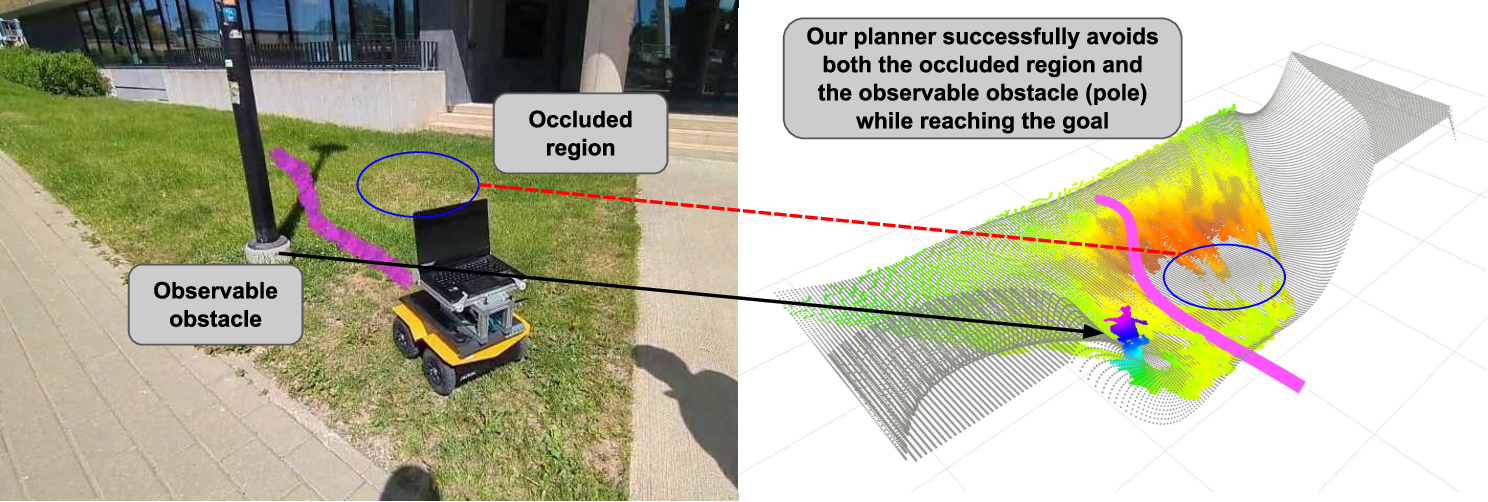}
    \caption{A wheeled robot navigating on off-road terrain. Safe path planning requires balancing a key trade-off: avoiding high-uncertainty regions caused by terrain-induced occlusions (highlighted in blue) while simultaneously steering clear of non-traversable areas, such as the pole on the left. The core idea behind our work is modeling terrain occlusion as coverage-induced epistemic uncertainty, which is propagated to the vehicle pose and contact points variable, and finally to the downstream planning cost. In this figure, the Grey surface represents our terrain reconstruction based on the partial point-cloud based observation of the terrain.}
    \label{fig:teaser_fig}
    \vspace{-0.5cm}
\end{figure}

Autonomous wheeled vehicles must traverse uneven terrain in applications such as forestry, construction, planetary exploration, agriculture, search and rescue, and off-road transportation. Unlike navigation on locally planar surfaces, off-road planning must account for the terrain-induced elevation, roll, pitch, and wheel contacts of the vehicle. A trajectory that appears favorable in the horizontal plane can place a wheel on a sharp elevation change, induce excessive body attitude, or lead to a near-tip-over configuration. For operations in which a quasi-static model is appropriate, stability-oriented trajectory planning therefore requires evaluating how each candidate planar trajectory interacts with the terrain.

Evaluating a trajectory requires computing a kinematic best-fit pose/contact-point configuration at successive positions. Given a planar position $(x_k,y_k)$ and yaw angle $\alpha_k$, at a given time step $k$, this model returns best-fit estimates of the terrain-induced elevation, roll, pitch, and candidate wheel--terrain contact locations. High-fidelity physics simulators can estimate these quantities accurately by modeling detailed contact dynamics~\cite{todorov2012mujoco, koenig2004design, agx}. However, they are generally too slow to serve as a predictive model within an online planning algorithm, which may evaluate thousands of such vehicle placements during each planning cycle. Prior work has therefore developed approximate but efficient pose and contact-point predictors for uneven terrain~\cite{singh2016feasible, xu2023efficient, manoharan2024bi}. In particular, Manoharan et al.~\cite{manoharan2024bi} formulate this prediction as a Nonlinear Least-Squares (NLS) problem. We adopt their model because it retains the kinematic coupling between the vehicle body, individual wheels, and terrain while avoiding full contact-dynamics simulation. However, like many model-based predictors, it treats the supplied terrain representation as deterministic.

Assuming a deterministic terrain representation is restrictive because the terrain is not measured at every location. Onboard depth sensors, such as LiDAR or RGB-D cameras, measure only surfaces that are visible along an unobstructed line of sight. On uneven terrain, a ridge, rock, tree, or other obstacle can block the view of the region behind it, leaving few or no depth measurements in that region. These unobserved areas appear as blind spots in the local map, as illustrated in Fig.~\ref{fig:teaser_fig}. Because planning requires a continuous terrain model, the missing elevations must be inferred from the surrounding visible measurements. This inference is not unique: several interpolated surfaces can fit the observed point cloud equally well while having substantially different elevations and slopes inside the blind spot. A reconstruction method may select the most likely surface as its nominal fit, but treating only this fit as the terrain model discards the variation among the other plausible interpolations. This discarded variation represents the epistemic uncertainty introduced by occlusion.

The importance of retaining this uncertainty becomes clear when evaluating wheel contacts. In a well-observed region, nearby measurements constrain the contact height and local surface normal. In a blind spot, each plausible interpolation can predict a different contact height and surface normal for the same wheel location. Passing these alternative terrain surfaces through the pose and contact-point predictor, therefore, produces different estimates of the vehicle's elevation, roll, pitch, and contact configuration. If the planner evaluates only the most likely terrain fit, it may predict a benign posture and accept a trajectory even though another plausible interpolation would induce a much larger tilt. Uncertainty-aware planning must therefore account not only for the nominal terrain fit, but also for how uncertainty in the unobserved terrain affects the predicted vehicle pose.

\noindent \textit{Core challenge:} Geometric traversability and sensing uncertainty must be considered jointly, but map uncertainty alone does not reveal its consequence for a particular vehicle trajectory. A large unobserved region away from the trajectory may have little effect on its predicted pose and contact costs, whereas a small blind spot beneath one wheel can substantially change the roll, pitch, or contact configuration. The relevant quantity is therefore not only where the terrain is uncertain, but how that uncertainty propagates through the wheel--terrain interaction model and affects the costs assigned to each candidate trajectory.

\noindent \textit{Approach:} We propose an occlusion-aware framework for generating quasi-static, stability-oriented reference trajectories from partial point-cloud observations. We represent the local terrain as a linear combination of fixed Fourier features and obtain the nominal coefficients $\boldsymbol{\lambda}^*$ by least-squares fitting. We then use the regularized inverse Hessian of the fitting loss as a terrain-parameter uncertainty estimate: parameter directions well constrained by the observed point cloud receive smaller uncertainty, whereas directions weakly constrained by occluded or sparsely observed regions receive larger uncertainty. Together with $\boldsymbol{\lambda}^*$, this matrix defines a local Gaussian surrogate used for first-order uncertainty propagation. It characterizes epistemic uncertainty caused by missing spatial support rather than a calibrated Bayesian posterior over sensor noise or model mismatch. The inverse-Hessian estimate is agnostic to the cause of missing observation support: terrain occlusion is the primary setting studied here, but range limits, field-of-view boundaries, and sparse sampling can produce the same weakly constrained parameter directions.

For each queried planar configuration $(x_k,y_k,\alpha_k)$, the NLS wheel--terrain model predicts the vehicle elevation, pitch, roll, and four wheel-contact locations. We use JAXopt~\cite{jaxopt_implicit_diff} implicit differentiation of the converged NLS optimality conditions to compute the sensitivity of this solution to the terrain parameters. Combining this sensitivity with the inverse-Hessian terrain uncertainty gives a first-order uncertainty estimate for the predicted pose and contacts. This makes the blind-spot uncertainty vehicle and trajectory-dependent rather than a generic map penalty.

We embed these estimates in a Cross-Entropy Method (CEM) trajectory optimizer whose objective combines curvature and acceleration penalties with uncertainty-aware pose and per-wheel surface-normal costs. The pose cost penalizes large predicted roll and pitch angles together with their propagated uncertainty. The surface-normal cost penalizes unfavorable terrain orientation at each predicted wheel contact and accounts for uncertainty through two pathways: directly through uncertainty in the terrain surface and indirectly through uncertainty in the terrain-dependent pose and contact locations. Consequently, CEM favors trajectories with smaller predicted attitude and surface-normal costs while keeping the wheel contacts in regions that are well constrained by the sensor observations. Regions that appear traversable under the nominal terrain fit but remain poorly observed receive a higher cost. This allows the planner to seek trajectories with lower modeled vehicle--terrain interaction costs in better-observed regions, while retaining model-based pose, contact, and uncertainty reasoning.

To improve computational efficiency, we use learning to warm-start the terrain-fitting stage of our pipeline while retaining model-based pose, contact, and uncertainty evaluation. We train a conditional Flow Matching (FM) model offline, which draws one Fourier-parameter prediction from the local Digital Elevation Model (DEM) to warm-start one to two Levenberg--Marquardt refinement iterations. In this way, learning reduces terrain-fitting latency, with the subsequent refinement iterations recovering the accuracy of the converged fit.

\noindent \textit{Contributions:} In summary, the main contributions are:
\begin{itemize}
    \item We fit partial terrain observations with fixed Fourier features and construct a regularized inverse-Hessian uncertainty estimate that exposes parameter directions weakly constrained by occluded and sparsely observed regions.

    \item Building on an NLS wheel--terrain interaction model, we compute its terrain-parameter sensitivity through implicit differentiation and combine it with the inverse-Hessian estimate to obtain first-order pose, contact-point, and per-wheel surface-normal uncertainty estimates. We incorporate these estimates into uncertainty-inflated costs for CEM-based trajectory optimization.

    \item We introduce an FM model that warm-starts terrain fitting and evaluate its fitting-accuracy--latency trade-off while retaining model-based refinement.
\end{itemize}

Across six synthetic terrains, every planner configuration was evaluated on the same 30 randomly sampled start--goal pairs per terrain. The proposed planner attained an aggregate failure rate of 18.9\%, compared with 46.1\% for Bi-level-opt~\cite{manoharan2024bi} and 41.7\% for GP-navigation~\cite{leininger2024gaussian}. These correspond to reductions of 27.2 and 22.8 percentage points, respectively. An ablation that sets $c_{\mathrm{unc}}=0$ has a failure rate of 34.4\%, 15.6 percentage points above the complete planner. Two illustrative hardware executions further demonstrate end-to-end planning near sensor blind spots, and the FM terrain-parameter network achieves a $14.3\times$ speedup over the converged nonlinear solver, with fitting RMSE increasing by 0.06\,mm ($3.1\%$).

\section{Related Work}

The literature most closely related to this work spans four coupled problems: predicting vehicle pose on uneven terrain, converting terrain geometry into traversability costs, incorporating terrain uncertainty into navigation, and using learning to estimate terrain safety or accelerate planning. The following subsections examine each direction and identify the corresponding gap addressed by the proposed framework.

\subsection{Pose Prediction for Vehicles on Uneven Terrain}

Evaluating a candidate trajectory on uneven terrain requires estimating a best-fit body attitude and candidate wheel-contact locations with an approximate interaction model. High-fidelity simulators can model detailed contact dynamics~\cite{todorov2012mujoco,koenig2004design,agx,benatti2022end}, but are often too costly for the thousands of repeated placement queries required by a sampling-based optimizer. Terrain-based approximations reduce this cost. Jordan and Zell~\cite{jordan2017real} represent wheels and terrain as elevation maps to identify contact points and feasible poses, whereas Fabian et al.~\cite{fabian2020pose} formulate pose prediction as image-processing operations between robot and terrain height maps. Jun et al.~\cite{jun2016pose} solve a complementarity-based contact problem for a tracked robot and use the resulting force-angle tip-over margin~\cite{papadopoulos2000force} during planning. More recent planners embed vehicle--terrain prediction directly into trajectory evaluation: UT-Planner~\cite{ning2026ut} estimates future attitude from four-wheel interactions with local point-cloud patches, and Xu et al.~\cite{xu2023efficient} incorporate terrain-dependent vehicle states into efficient trajectory planning. The closest nominal interaction model to ours is the differentiable NLS formulation of Manoharan et al.~\cite{manoharan2024bi}, which predicts vehicle pose and wheel contacts and is embedded in a bi-level trajectory optimizer. 


\begin{contribbox}
Relative to deterministic pose/contact prediction \cite{ning2026ut}, \cite{xu2023efficient}, including the NLS predictor shared with the bi-level method~\cite{manoharan2024bi}, we treat the fitted terrain as uncertain. Implicit differentiation of the converged NLS optimality conditions maps the regularized inverse-Hessian terrain estimate into first-order uncertainty estimates for the vehicle pose and wheel contacts, which subsequently affect the trajectory cost.
\end{contribbox}

\subsection{Traversability Analysis}
\noindent
A broad class of traversability methods first summarizes the
terrain around a map cell, path node, or vehicle-sized patch by
a scalar cost and then supplies that cost to a planner. Waibel
et al.~\cite{waibel2022rough} construct continuous local and
global cost maps using either patch-scale vertical dispersion
and fitted-plane inclination or a learned aggregate inertial
response. These costs rank the map cells and trajectories generated
by a planar Ackermann model, without explicitly solving for the
terrain-induced body pose or individual wheel contacts. Jian et
al.~\cite{jian2022putn} similarly fit a local plane around each
node generated by a sampling-based global planner and combine
its slope, flatness, and point-cloud sparsity into a scalar
traversability index. Gaussian-process regression interpolates
this index and its predictive variance along the resulting path,
which a nonlinear model predictive controller uses to regulate
the vehicle's motion. Although the fitted plane defines a local
surface frame, it is not a four-wheel pose/contact solution.

Gaussian-process methods improve the continuity and temporal
consistency of this map-level representation. Tan et
al.~\cite{tan2025real} use a sparse Gaussian process to predict
terrain elevation from point-cloud curvature and height gradient
features. Curvature, gradient, and slope are then combined into
a per-cell traversability score, while predictive variance
influences the fusion of current and historical map estimates
before graph-based path search and trajectory smoothing. Wang
et al.~\cite{Wang_2025} use Bayesian generalized-kernel
inference to estimate unobserved terrain attributes and construct
continuous traversability and uncertainty maps for the global path
search and predictive motion control. In both cases, the planner
consumes a spatial traversability field: uncertainty remains
associated with the inferred map or its scalar score rather than
being propagated through a vehicle--terrain interaction model.

Other methods introduce vehicle information without estimating
the four-contact configuration considered here. Tian et
al.~\cite{tian2023efficient} use elevation-map normals, terrain
roughness, and vehicle kinematics during graph search to expand
terrain-conforming path nodes and evaluate a terrain-based
traversal cost. Yu et al.~\cite{yu2026real} go further by
incorporating a known height field into a three-dimensional
dynamic vehicle model. Local surface derivatives determine
terrain-induced roll and pitch, while gravity, slip, and load
transfer, friction limits, and tire-lift constraints enter a
unified predictive planning-and-control formulation. Their
nominal model, however, assumes that the terrain surface is known
and that the vehicle remains tangent to it; it does not reconstruct
a body pose and four wheel-contact locations from partial
observations or propagate terrain-reconstruction uncertainty
through those quantities.

\begin{contribbox}
The proposed method differs in the quantity evaluated by the
planner. Rather than assigning one precomputed traversability
value to a cell, patch, or path node, every candidate vehicle
configuration is passed through a nonlinear least-squares
wheel--terrain model that estimates the terrain-induced body
elevation, roll, pitch, and four candidate contact locations.
Implicit differentiation then propagates the inverse-Hessian
terrain uncertainty through this solution and through the
surface normals evaluated at the predicted contacts.
Consequently, terrain geometry and observation support are
penalized according to their estimated effect on a particular
chassis placement and its four wheel locations, rather than
solely through a location-level slope, roughness, or
traversability score.
\end{contribbox}

\subsection{Uncertainty-Aware Navigation}

Risk-aware off-road planners differ not only in the source
of uncertainty, but also in how that is used in the downstream
planner. STEP~\cite{fan2021step} constructs
uncertainty-aware risk layers for collision, step height,
tip-over, contact loss, and slippage. The mean and variance
assigned to these risk factors are combined into a
conditional value-at-risk traversability cost used by
grid-based geometric planning and receding-horizon
kinodynamic planning. STEP represents tip-over and contact
loss as map-level risk scores. Tip-over is evaluated from
local terrain slope and vehicle orientation, while contact
loss is inferred from plane-fitting residuals. Uncertainty is
assigned directly to these risk scores; STEP does not estimate
uncertainty in a terrain-induced three-dimensional chassis pose
or four individual wheel-contact locations. In contrast, our
method solves for these vehicle-dependent quantities at each
candidate configuration and propagates terrain uncertainty
through that solution.

Yin et al.~\cite{yin2023reliable} instead propagate
uncertainty in terrain and vehicle properties through
physics-based mobility simulations and surrogate models.
Reliability constraints on maximum attainable speed and vertical acceleration are then imposed during the global path search. This is model-based uncertainty propagation, but its
outputs are mobility-failure probabilities rather than
uncertainty in the terrain-induced chassis attitude, contact
locations, or contact-surface normals. Hu et
al.~\cite{hu2026safe} combine distributional reinforcement
learning and conditional value-at-risk for route selection
with a terramechanics-based adaptive tracking controller that
compensates for wheel slip, modeling error, and external
disturbances. Their planning and control stages address the route
risk and tracking robustness, respectively, without propagating
terrain-reconstruction uncertainty into the vehicle pose and
wheel contacts used to evaluate a candidate configuration.

Probabilistic mapping and active exploration methods retain
uncertainty further upstream. PTS-Map~\cite{kim2024pts}
sequentially estimates probabilistic ground-surface and
above-ground elevation states, using ground-state uncertainty
to reduce the influence of unreliable geometric measurements
and construct a temporally consistent traversability map.
Thus, its uncertainty describes the terrain-map state rather
than the distribution of chassis attitude and contacts induced
by placing four separated wheels on that terrain. Beyer et
al.~\cite{beyer2024risk} predict how a learned risk map will
evolve with the path and speed of the vehicle and bound the
expected future risk of candidate trajectories. Confidence-aware
exploration~\cite{park2025cute}, joint mapping and planning in
belief space~\cite{pairet2021online}, and slope-aware informative
sampling~\cite{tazaki2025slope} similarly uses uncertainty to
choose observations, reject poorly supported regions, or
enforce planning-level safety bounds. They do not propagate
terrain-reconstruction uncertainty through a wheel--terrain
pose/contact solution.

The Gaussian-process terrain-mapless method of Leininger et
al.~\cite{leininger2024gaussian} is closest to ours in its
treatment of sparsely observed elevation and therefore serves
as an experimental baseline. It predicts elevation and
geometric traversability from a sparse Gaussian process and
marks cells as non-traversable when the elevation variance
exceeds a prescribed threshold. This mask identifies poorly
supported map regions, but it assigns the same map-level
interpretation of that uncertainty, regardless of how a
candidate chassis places its four wheels relative to the
region.

\begin{contribbox}
Unlike methods that pass a map-level risk or confidence value
to the planner, the proposed framework evaluates how terrain
uncertainty affects a specific vehicle configuration. Starting
from a regularized inverse-Hessian uncertainty estimate for the
fitted terrain parameters, implicit differentiation propagates
this uncertainty through the nonlinear least-squares solution
for body pose and four wheel contacts. This produces first-order
uncertainty estimates for vehicle attitude, contact locations,
and surface normals at the predicted contacts, which are
incorporated into the trajectory objective. An uncertain terrain
region is thus penalized according to its predicted effect on a
particular chassis placement and its four wheel contacts, rather
than according to a precomputed risk, confidence, or
mobility-reliability value.
\end{contribbox}

\subsection{Learning-Based Methods}

Learning-based uneven-terrain navigation spans traversability estimation, control, and computational acceleration; Bhosale et al.~\cite{bhosale2026review} provide a broader review. Existing methods use learning for predicting vehicle attitude~\cite{datar2024learning}, Deep Reinforcement Learning (DRL) for rough-terrain control~\cite{wiberg2021control}, self-supervision and DRL for traversability costs~\cite{castro2023does,weerakoon2022terp}, and camera--LiDAR data for uncertain elevation and traversability prediction~\cite{frey2024roadrunner}. Others learn uncertainty-aware policies~\cite{lee2023learning}, feature-level uncertainty~\cite{triest2024unrealnet}, or probabilistic traversability with test-time adaptation~\cite{endo2026deep}. Their uncertainty remains at the perception, traversability, or policy output rather than being propagated through a three-dimensional pose and wheel-contact solution.

Learning can also amortize model-based computation. Neural surrogates replace repeated vehicle--terrain simulations for mobility prediction~\cite{hua2024double}, while Motion Planning Diffusion~\cite{carvalho2023motion} and FlowMP~\cite{nguyen2025flowmp} learn trajectory priors or generators. In our framework, Flow Matching has a narrower role: it only initializes terrain fitting. Levenberg--Marquardt still refines the terrain parameters against the current point cloud.

\begin{contribbox}
Our Flow Matching model accelerates terrain-fit initialization without replacing terrain estimation. The current point cloud determines the refined surface and its uncertainty. CEM is unchanged and still ranks candidate trajectories based on predicted body pose, four-wheel contacts, and contact-normal uncertainty. Learning, therefore, improves computational efficiency while the safety-relevant reasoning remains model-based.
\end{contribbox}

\input{methods_3_iiib_likelihood}

\input{Results.tex}

\input{Conclusions.tex}

\section*{Acknowledgments}

During the preparation of this manuscript, the authors used generative AI tools for editorial support, including proofreading, improving readability, and restructuring prose. The authors reviewed and revised all resulting text. The tools did not contribute to the underlying research: the authors independently developed the methods, derived the mathematics, designed and conducted the experiments, interpreted the results, and prepared the reported figures and data. The authors retain full responsibility for the manuscript's technical accuracy and conclusions.

\IEEEtriggeratref{44}
\bibliographystyle{IEEEtran}
\bibliography{references}

\end{document}

%% file: methods_3_iiib_likelihood.tex
\section{Methods}

\subsubsection*{Symbols and Notation}
\noindent The transpose of a vector or matrix is denoted by the superscript $\mathsf{T}$. Matrices are represented using uppercase bold letters, while lowercase italic letters are used for scalars. Vectors are denoted by bold lowercase letters (upright for Latin symbols and italic for Greek symbols). Variance is denoted by $\operatorname{Var}[\cdot]$, and the determinant of a matrix is represented by $\det(\cdot)$. The time index is denoted by the subscript $k$. Important symbols are defined in Table~\ref{tab:symbols_notation}.

\begin{table}[t]
    \centering
    \caption{Important symbols.}
    \label{tab:symbols_notation}
    \footnotesize
    \renewcommand{\arraystretch}{1.1}
    \setlength{\tabcolsep}{4pt}
    \begin{tabular}{@{}p{0.30\columnwidth}p{0.66\columnwidth}@{}}
        \toprule
        \textbf{Symbol} & \textbf{Description} \\
        \midrule
        $\mathbf{x}_k$ & Planar query state $(x_k,y_k,\alpha_k)$ at time step $k$. \\
        $\mathbf{p}_k$ & Planar position $(x_k,y_k)$ at time step $k$. \\
        $\boldsymbol{\xi}_k$, $\boldsymbol{\xi}_k^*$ & Terrain-dependent best-fit state: elevation, pitch, roll, and candidate wheel--terrain contact points, and its converged least-squares estimate. \\
        $f_{\boldsymbol{\lambda}}(x,y)$ & Terrain elevation model parameterized by Fourier coefficients. \\
        $\boldsymbol{\lambda},\,\boldsymbol{\lambda}^*$ & Terrain parameter vector and its converged least-squares estimate. \\
        $\boldsymbol{\Sigma}_{\lambda}$ & Regularized inverse-Hessian terrain-parameter uncertainty estimate. \\
        $\boldsymbol{\phi}(x,y)$ & Fourier feature vector at spatial query $(x,y)$. \\
        $\boldsymbol{\Phi}$ & Terrain-regression design matrix. \\
        $\mathbf{g}$ & Stacked residual vector of wheel--terrain kinematic constraints. \\
        $\mathbf{J}_{\boldsymbol{\xi}}$ & Jacobian of $\mathbf{g}$ with respect to $\boldsymbol{\xi}_k$. \\
        $\mathbf{J}_{\boldsymbol{\lambda}}$ & Jacobian of $\mathbf{g}$ with respect to $\boldsymbol{\lambda}$. \\
        $\mathbf{J}_{\boldsymbol{\lambda},k}^{\boldsymbol{\xi}}$ & Implicit sensitivity $\mathrm{d}\boldsymbol{\xi}_k^*/\mathrm{d}\boldsymbol{\lambda}$. \\
        $\boldsymbol{\Sigma}_{\xi_k}$ & First-order propagated uncertainty estimate for pose and contact points. \\
        $H$ & Number of equal-duration future intervals and future trajectory samples. \\
        $\Delta t,\,T$ & Planning interval and horizon duration, with $T=H\Delta t$. \\
        $M$ & Number of retained point-cloud samples used for terrain fitting. \\
        $N$ & Number of Fourier frequencies in the terrain representation. \\
        $\mathbf{W},\dot{\mathbf{W}},\ddot{\mathbf{W}}$ & Bernstein basis matrix and its physical-time first and second derivatives. \\
        $n_c$ & Number of polynomial coefficients per trajectory dimension. \\
        $n_{cem},\,n_s,\,n_e$ & Number of CEM iterations, sampled trajectories, and elite trajectories. \\
        $c_{cem}$ & Total CEM objective, comprising nominal and uncertainty terms. \\
        $c_{\kappa},\,c_a,\,c_n,\,c_p$ & Curvature, acceleration, surface-normal, and pose costs. \\
        $w_{\kappa},\,w_a,\,w_n,\,w_p$ & Weights for the curvature, acceleration, normal, and pose costs. \\
        $\rho_n,\,\rho_p$ & Penalty coefficients for uncertainty-aware costs. \\
        \bottomrule
    \end{tabular}
\end{table}

\subsection{Method Overview}
\begin{figure*} 
    \centering
    \begin{subfigure}{\linewidth} 
        \centering
        \includegraphics[scale=0.8]{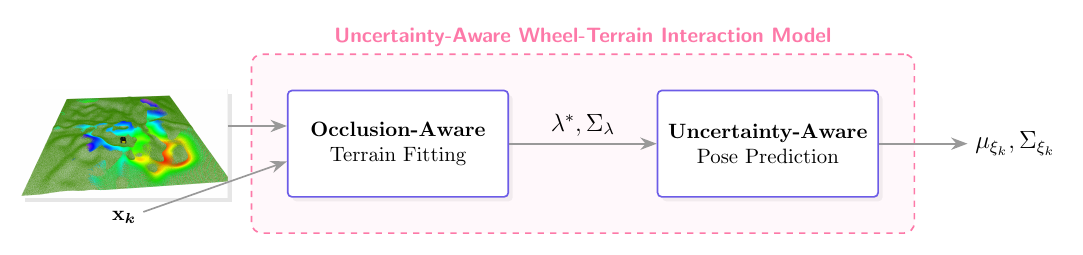}
        \caption{Uncertainty-aware wheel--terrain interaction model}
        \label{fig:terrain_pipeline}
    \end{subfigure}
    
    \vspace{0.5em}
    
    \begin{subfigure}{\linewidth}
        \centering
        \includegraphics[scale=0.8]{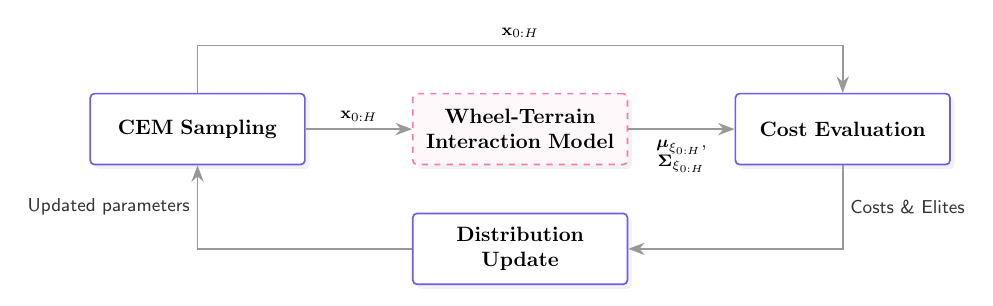}
        \caption{Model-based CEM trajectory-optimization pipeline}
        \label{fig:cem_pipeline}
    \end{subfigure}
    
    \caption[Model-based core of the proposed framework.]{Model-based core of the proposed framework. (a) Partial terrain observations are converted into uncertain terrain parameters and propagated through the pose/contact model. (b) CEM evaluates sampled trajectories using the resulting nominal states and uncertainty. The Flow Matching module introduced subsequently provides a warm start for terrain refinement without affecting the model-based evaluation shown here.}
    \label{fig:combined_pipelines}
\end{figure*}

\noindent Fig.~\ref{fig:combined_pipelines} presents the model-based core of our approach. We consider a rigid, non-articulated four-wheeled vehicle operating on terrain represented as a single-valued elevation field. The vehicle is assumed to move slowly enough for a quasi-static model to be appropriate; slip and dynamic contact forces are not modeled. A candidate trajectory is represented by a sequence of planar query states. At time step $k$, the query state contains the planned position $(x_k,y_k)$ and yaw angle $\alpha_k$:
\begin{equation}
    \mathbf{x}_k = 
    \begin{bmatrix}
        x_k & y_k & \alpha_k
    \end{bmatrix}^{\mathsf{T}}.
\end{equation}
Given $\mathbf{x}_k$ and a terrain surface, the wheel--terrain interaction model returns a kinematic best-fit estimate of the body elevation $z_k$, pitch $\beta_k$, roll $\gamma_k$, and candidate 3D contact point $(x_{c_i,k},y_{c_i,k},z_{c_i,k})$ of each wheel $i$. We collect these quantities into the terrain-dependent state $\boldsymbol{\xi}_k\in\mathbb{R}^{15}$.

The upper pipeline in Fig.~\ref{fig:combined_pipelines} shows how uncertainty in the partial terrain observation is transferred to this state. Least-squares terrain fitting provides the nominal Fourier parameters $\boldsymbol{\lambda}^*$, while the regularized inverse Hessian of the fitting loss provides the parameter-uncertainty estimate $\boldsymbol{\Sigma}_{\lambda}$. We use $\mathcal{N}(\boldsymbol{\lambda}^*,\boldsymbol{\Sigma}_{\lambda})$ as a local Gaussian surrogate for coverage-induced epistemic uncertainty, not as an exact Bayesian posterior. For a query $\mathbf{x}_k$, the NLS best-fit pose/contact-point solver evaluated at $\boldsymbol{\lambda}^*$ provides the nominal state $\boldsymbol{\xi}_k^*$. We use JAXopt implicit differentiation of the converged NLS optimality conditions and first-order uncertainty propagation to obtain
\begin{equation}
    \boldsymbol{\mu}_{\xi_k}=\boldsymbol{\xi}_k^*,
    \qquad
    \boldsymbol{\Sigma}_{\xi_k}\approx
    \mathbf{J}_{\boldsymbol{\lambda},k}^{\boldsymbol{\xi}}
    \boldsymbol{\Sigma}_{\lambda}
    \left(\mathbf{J}_{\boldsymbol{\lambda},k}^{\boldsymbol{\xi}}\right)^{\mathsf{T}}.
\end{equation}
Thus, the terrain-dependent state is represented by a nominal prediction together with a first-order uncertainty estimate that quantifies how strongly that prediction depends on uncertain terrain geometry.

The lower pipeline embeds this interaction model in a sampling-based trajectory optimizer based on CEM~\cite{rubinstein1999cross}. Starting from the measured planar state $\mathbf{p}_0$, CEM samples the $H$ future positions $\{\mathbf{p}_k\}_{k=1}^{H}$ through polynomial coefficients, queries the interaction model along each trajectory, and evaluates curvature and acceleration together with uncertainty-aware pose and per-wheel surface-normal costs. The pose and normal terms account for nominal attitude and terrain-normal geometry together with the uncertainty induced by incomplete terrain observations. CEM then updates its sampling distribution from the elite candidates, steering the search toward trajectories with lower predicted attitude and surface-normal costs in better-observed regions.

To reduce the terrain-fitting latency, a conditional Flow Matching model is trained offline and attached to this model-based core. The terrain-parameter model draws one prediction from the local elevation map to initialize the refinement of $\boldsymbol{\lambda}^*$. This learned module warm-starts only the terrain refinement and leaves the downstream CEM planner unchanged. The following subsections develop the model-based components in Fig.~\ref{fig:combined_pipelines}; the learned warm-start module is described subsequently.

\subsection{Uncertainty-Aware Representation of the Terrain}
\label{sec:terrain_model}

\noindent We first construct a continuous local terrain representation from the discrete elevation samples in the point cloud. Let
\begin{equation}
    \mathcal{D}=\{(x^j,y^j,z^j)\}_{j=1}^{M}
\end{equation}
denote the retained samples from a single input point-cloud frame. Before fitting, the cloud is voxel-filtered and, when a point cap is required, randomly sampled without replacement. All retained samples are transformed into the vehicle frame, whose origin is the vehicle reference point, with $x$, $y$, and $z$ measured in meters. Let $\mathbf{p}=[x\;y]^{\mathsf{T}}$ and $\mathbf{p}^j=[x^j\;y^j]^{\mathsf{T}}$. We fit a deterministic surface $f_{\boldsymbol{\lambda}}(x,y)$ to the observations and use their spatial support to quantify where that fit is weakly constrained. The surface is represented using fixed Fourier features:
\begin{equation}
\begin{aligned}
    f_{\boldsymbol{\lambda}}(\mathbf{p})
    = \sum_{n=1}^{N}
    &a_n\cos\!\left((\boldsymbol{\omega}^{c}_{n})^{\mathsf{T}}\mathbf{p}\right) \\
    &+b_n\sin\!\left((\boldsymbol{\omega}^{s}_{n})^{\mathsf{T}}\mathbf{p}\right),
\end{aligned}
\label{eqn:fourier_terrain}
\end{equation}
where the cosine and sine frequencies are sampled independently,
\begin{equation}
    \boldsymbol{\omega}^{c}_{n},\boldsymbol{\omega}^{s}_{n}
    \overset{\mathrm{i.i.d.}}{\sim}
    \mathcal{N}(\mathbf{0},\mathbf{I}_2),
\end{equation}
and are then held fixed across all training and evaluation terrain maps, defining a shared random Fourier basis. For each terrain, the fitted amplitudes select and combine these fixed modes. Because $\mathbf{p}$ is expressed in meters, the frequencies have units of $\mathrm{m}^{-1}$. We use $N=100$ frequencies in all experiments.

The amplitudes are collected in
\begin{equation}
    \boldsymbol{\lambda}
    =
    \begin{bmatrix}
        a_1 & b_1 & \cdots & a_N & b_N
    \end{bmatrix}^{\mathsf{T}}
    \in\mathbb{R}^{2N}.
\end{equation}
Defining the corresponding feature vector
\begin{equation}
\begin{aligned}
    \boldsymbol{\phi}(\mathbf{p})
    =
    \begin{bmatrix}
        \cos((\boldsymbol{\omega}^{c}_{1})^{\mathsf{T}}\mathbf{p}) \\
        \sin((\boldsymbol{\omega}^{s}_{1})^{\mathsf{T}}\mathbf{p}) \\
        \vdots \\
        \cos((\boldsymbol{\omega}^{c}_{N})^{\mathsf{T}}\mathbf{p}) \\
        \sin((\boldsymbol{\omega}^{s}_{N})^{\mathsf{T}}\mathbf{p})
    \end{bmatrix},
\end{aligned}
\end{equation}
we can write
\begin{equation}
    f_{\boldsymbol{\lambda}}(\mathbf{p})
    =\boldsymbol{\phi}(\mathbf{p})^{\mathsf{T}}\boldsymbol{\lambda}.
\end{equation}
The model is therefore linear in $\boldsymbol{\lambda}$ even though it represents a nonlinear function of the spatial coordinates.

Stacking the observations gives
\begin{equation}
    \mathbf{z}
    =
    \begin{bmatrix}
        z^1 & \cdots & z^M
    \end{bmatrix}^{\mathsf{T}},
    \qquad
    \boldsymbol{\Phi}
    =
    \begin{bmatrix}
        \boldsymbol{\phi}(\mathbf{p}^1)^{\mathsf{T}} \\
        \vdots \\
        \boldsymbol{\phi}(\mathbf{p}^M)^{\mathsf{T}}
    \end{bmatrix},
\end{equation}
and we introduce the fixed homoscedastic Gaussian observation model
\begin{equation}
    \mathbf{z}\mid\boldsymbol{\lambda}
    \sim
    \mathcal{N}\!\left(
        \boldsymbol{\Phi}\boldsymbol{\lambda},
        \sigma_z^2\mathbf{I}_M
    \right).
    \label{eqn:terrain_likelihood}
\end{equation}
Here, $\sigma_z>0$ is a fixed terrain-elevation observation scale. We select it once from the empirical per-observation residuals of the converged terrain fits and hold it constant across all maps and experiments. The converged fitting RMSE is consistently of order $10^{-3}\,\mathrm{m}$, supporting the selected scale.

Dropping terms that are constant with respect to $\boldsymbol{\lambda}$, the negative log-likelihood associated with \eqref{eqn:terrain_likelihood} gives
\begin{equation}
\begin{aligned}
    \boldsymbol{\lambda}^{*}
    &=\arg\min_{\boldsymbol{\lambda}}
    \mathcal{L}_{\mathrm{terrain}}(\boldsymbol{\lambda}), \\
    \mathcal{L}_{\mathrm{terrain}}(\boldsymbol{\lambda})
    &=\frac{1}{2\sigma_z^2}
    \left\|
        \boldsymbol{\Phi}\boldsymbol{\lambda}-\mathbf{z}
    \right\|_2^2.
\end{aligned}
\label{eqn:terrain_fit}
\end{equation}
Because $\sigma_z$ is fixed, $\boldsymbol{\lambda}^{*}$ coincides with the ordinary least-squares solution. The scaling determines the curvature magnitude used to construct the uncertainty estimate below.

Each planning instance can contain up to $35{,}000$ retained observations, making repeated full-data regression a significant part of the online computation. We evaluate the pointwise residuals and solve the problem using the Levenberg--Marquardt implementation in JAXopt. During deployment, the Flow Matching prediction introduced in Section~\ref{flow_acc} warm-starts this solve, and one to two refinement iterations were empirically sufficient to recover the converged terrain parameters $\boldsymbol{\lambda}^{*}$.

The fitted surface is well constrained where the point cloud provides dense spatial support and weakly constrained where observations are missing. We quantify the resulting terrain-parameter uncertainty using the local curvature of the negative log-likelihood. Let
\begin{equation}
\begin{aligned}
    \mathbf{H}_{\boldsymbol{\lambda}}
    &=
    \left.
    \nabla_{\boldsymbol{\lambda}}^2
    \mathcal{L}_{\mathrm{terrain}}(\boldsymbol{\lambda})
    \right|_{\boldsymbol{\lambda}^{*}} \\
    &=\frac{1}{\sigma_z^2}
    \boldsymbol{\Phi}^{\mathsf{T}}\boldsymbol{\Phi}.
\end{aligned}
\label{eqn:terrain_hessian}
\end{equation}
The second equality follows from the fixed-feature linear model and the Gaussian likelihood. In implementation, the curvature can equivalently be evaluated through automatic differentiation or from the normal matrix $\boldsymbol{\Phi}^{\mathsf{T}}\boldsymbol{\Phi}$. We define the regularized inverse-Hessian parameter covariance
\begin{equation}
\begin{aligned}
    \boldsymbol{\Sigma}_{\lambda}
    &=
    \left(
        \mathbf{H}_{\boldsymbol{\lambda}}
        +\frac{\eta}{\sigma_z^2}\mathbf{I}_{2N}
    \right)^{-1} \\
    &=
    \sigma_z^2
    \left(
        \boldsymbol{\Phi}^{\mathsf{T}}\boldsymbol{\Phi}
        {}+\eta\mathbf{I}_{2N}
    \right)^{-1},
    \qquad
    \eta=10^{-3}.
\end{aligned}
\label{eqn:sigma_lambda}
\end{equation}
The fixed damping value $\eta=10^{-3}$ stabilizes parameter directions with little curvature. It is applied when constructing $\boldsymbol{\Sigma}_{\lambda}$, while the nominal parameter vector remains the likelihood solution defined in \eqref{eqn:terrain_fit}. The factor $\sigma_z^2$ gives the uncertainty estimate of the physical scale of squared terrain elevation.

For the summed loss in \eqref{eqn:terrain_fit}, each distinct retained observation contributes
$\sigma_z^{-2}\boldsymbol{\phi}(\mathbf{p}^j)\boldsymbol{\phi}(\mathbf{p}^j)^{\mathsf{T}}$
to the Hessian. Sampling without replacement ensures that every retained point contributes once. Parameter directions supported by many observations therefore have large curvature and small inverse curvature, whereas directions weakly constrained by the current view retain larger uncertainty. If $h_i$ is an eigenvalue of $\boldsymbol{\Phi}^{\mathsf{T}}\boldsymbol{\Phi}$, the corresponding parameter variance is $\sigma_z^2/(h_i+\eta)$. Thus, $\sigma_z$ sets the common observation scale, while the spatial distribution of the current point cloud determines how strongly uncertainty is amplified in poorly observed directions.

\begin{figure*}
    \centering
    \begin{subfigure}{0.3\linewidth}
        \includegraphics[width=\linewidth, height=3.5cm]{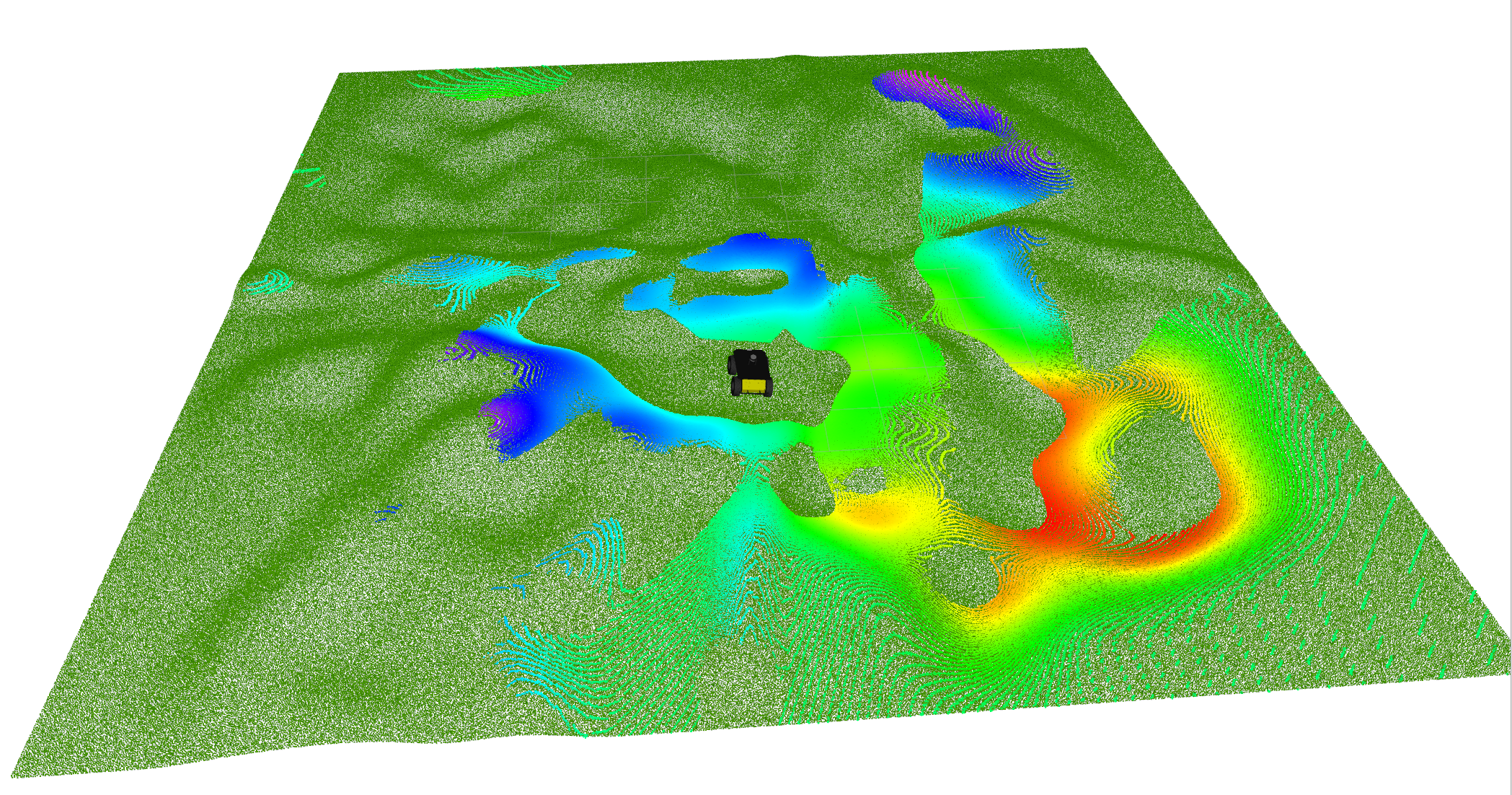}
        \caption{}
        \label{fig:occlusion_scenario}
    \end{subfigure}
    \hspace{1mm}
    \begin{subfigure}{0.3\linewidth}
        \includegraphics[width=\linewidth, height=3.5cm]{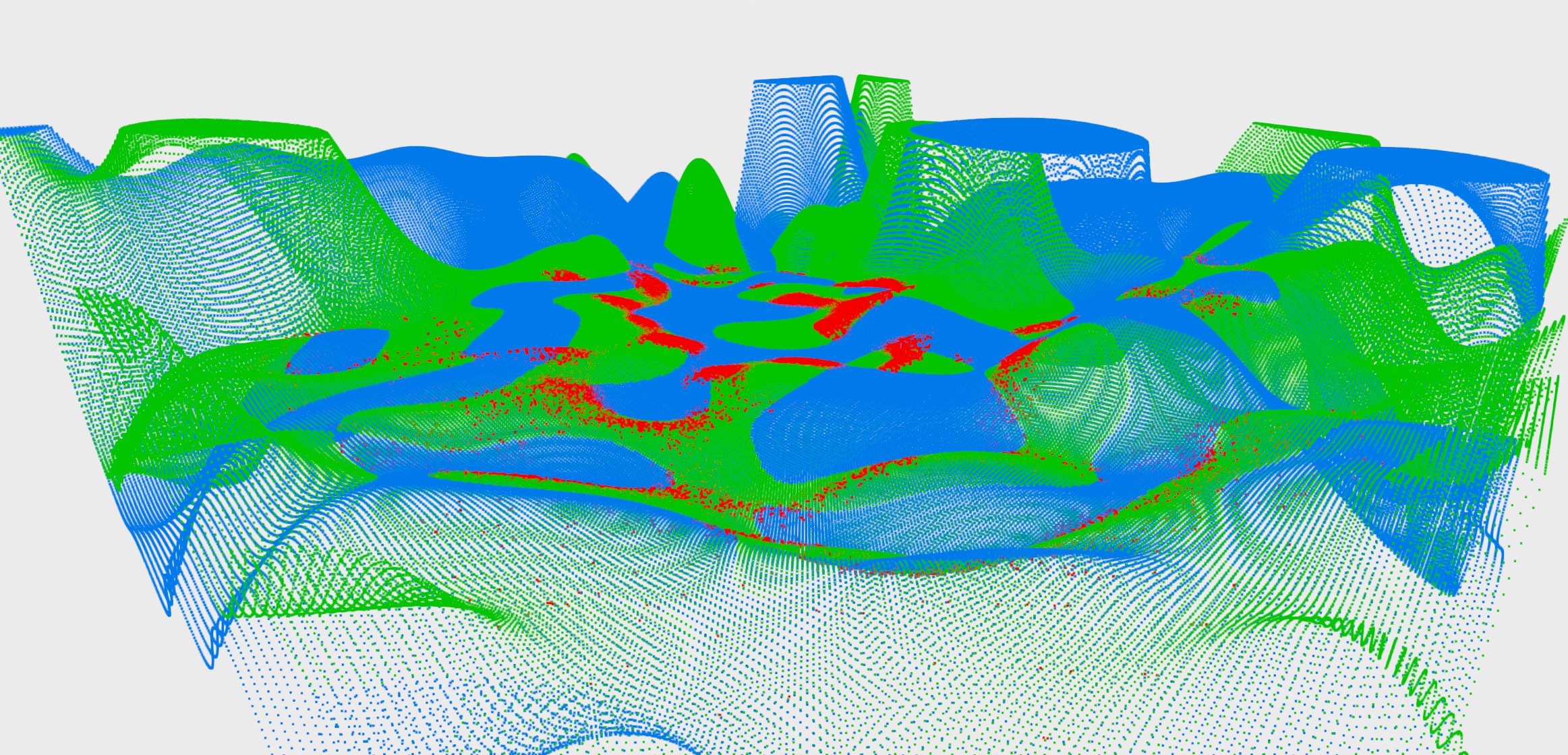}
        \caption{}
        \label{fig:terrain_fit_occlusion}
    \end{subfigure}
    \hspace{1mm}
    \begin{subfigure}{0.3\linewidth}
        \includegraphics[width=\linewidth]{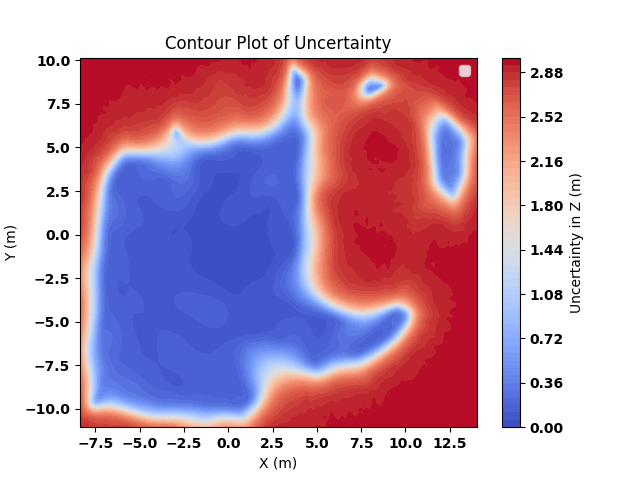}
        \caption{}
        \label{fig:elevation_uncert}
    \end{subfigure}
    \caption{Likelihood-scaled terrain uncertainty from partial observations. (a) On-board sensing may fail to observe parts of the terrain because of occlusion. (b) The nominal Fourier fit can be unreliable where spatial observations are sparse or missing. Two sample fits are shown (blue and green), matching at points where data are available (red) and diverging in unseen areas. (c) The regularized inverse-Hessian variance is larger in directions that are weakly constrained by the observed point cloud.}
    \label{fig:occlusion_description}
\end{figure*}

At a query point $\mathbf{p}=[x\;y]^{\mathsf{T}}$, the corresponding latent terrain-height variance is
\begin{equation}
    u_f(\mathbf{p})
    =
    \boldsymbol{\phi}(\mathbf{p})^{\mathsf{T}}
    \boldsymbol{\Sigma}_{\lambda}
    \boldsymbol{\phi}(\mathbf{p}).
    \label{eqn:terrain_height_variance}
\end{equation}
The quantity $u_f(\mathbf{p})$ is the estimated variance of the latent terrain height at $\mathbf{p}$. It is generally smaller in densely observed regions and larger in occluded or sparsely observed regions, as illustrated in Fig.~\ref{fig:occlusion_description}. A future noisy elevation measurement would additionally contain the observation variance $\sigma_z^2$, giving $u_f(\mathbf{p})+\sigma_z^2$. The online planner uses $\boldsymbol{\Sigma}_{\lambda}$ directly and propagates it analytically through the vehicle model in the following subsection. Samples from $\mathcal{N}(\boldsymbol{\lambda}^{*},\boldsymbol{\Sigma}_{\lambda})$ are used to visualize the terrain variations represented by this local surrogate.

This construction provides a local, single-frame uncertainty estimate determined by the spatial support of the current observations. Temporal belief updates, correlated sensor errors, and terrain-model mismatch beyond the local inverse-Hessian approximation are outside the present formulation.

\subsection{Uncertainty-Aware Pose and Contact-Point Prediction}
\label{sec:pose_prediction}

\noindent We next propagate terrain uncertainty into uncertainty over the vehicle pose and wheel--terrain contact points. The controllable query is the yaw-plane state
\begin{equation}
    \mathbf{x}_k
    =
    \begin{bmatrix}
        x_k & y_k & \alpha_k
    \end{bmatrix}^{\mathsf{T}}.
\end{equation}
The terrain-dependent variables are the body elevation $z_k$, pitch $\beta_k$, roll $\gamma_k$, and four candidate 3D wheel--terrain contact-point variables. We define
\begin{equation}
    \boldsymbol{\xi}_{k}
    =
    \begin{bmatrix}
        z_k &
        \beta_k &
        \gamma_k &
        \mathbf{p}_{oc,1}^{\mathsf{T}} &
        \mathbf{p}_{oc,2}^{\mathsf{T}} &
        \mathbf{p}_{oc,3}^{\mathsf{T}} &
        \mathbf{p}_{oc,4}^{\mathsf{T}}
    \end{bmatrix}^{\mathsf{T}}
    \in \mathbb{R}^{15},
    \label{eqn:xi_def}
\end{equation}
where
\begin{equation}
    \mathbf{p}_{oc,i}
    =
    \begin{bmatrix}
        x_{c_i,k} & y_{c_i,k} & z_{c_i,k}
    \end{bmatrix}^{\mathsf{T}}
\end{equation}
is the candidate global contact-point variable associated with wheel $i$.

Following the parallel-manipulator interpretation used in prior terrain-pose prediction models~\cite{chakraborty2004kinematics,singh2016feasible,manoharan2024bi}, we define a global inertial frame $\{O\}$, a body-fixed frame $\{L\}$, and a translating frame $\{G\}$ that shares the body origin but keeps the orientation of the inertial frame. Let $\mathbf{R}(\alpha_k,\beta_k,\gamma_k)$ be the rotation from $\{L\}$ to $\{G\}$. Under the fixed-angle convention, $\alpha_k$ is yaw about the $z$-axis, $\beta_k$ is pitch about the $y$-axis, and $\gamma_k$ is roll about the $x$-axis, such that $\mathbf{R}(\alpha_k,\beta_k,\gamma_k)=\mathbf{R}_z(\alpha_k)\mathbf{R}_y(\beta_k)\mathbf{R}_x(\gamma_k)$. The rigid chassis geometry defines a three-component loop-closure residual for each wheel:
\begin{align}
    \mathbf{r}^{\mathrm{kin}}_{k,i}
    &=
    \mathbf{p}_{og}
    +
    \mathbf{p}_{gc,i}
    -
    \mathbf{p}_{oc,i}
    \in\mathbb{R}^{3},
    \label{eqn:pos_vec}
    \\
    \mathbf{p}_{gc,i}
    &=
    \mathbf{R}(\alpha_k,\beta_k,\gamma_k)
    \begin{bmatrix}
        \delta_i l \\
        r_i w \\
        -h_i
    \end{bmatrix},
    \quad
    i\in\{1,2,3,4\},
    \label{eqn:pgci}
    \\
    \delta_i
    &=
    \begin{cases}
    1, & i\in\{1,4\},\\
    -1, & i\in\{2,3\},
    \end{cases}
    \qquad
    r_i
    =
    \frac{2.5-i}{|2.5-i|},
    \\
    \mathbf{p}_{og}
    &=
    \begin{bmatrix}
        x_k & y_k & z_k
    \end{bmatrix}^{\mathsf{T}}.
    \label{eqn:pog}
\end{align}
Here $l$ and $w$ are the half-length and half-width of the chassis, and $h_i$ denotes the effective vertical offset from the body frame to the wheel contact point, including the wheel radius.

Each candidate contact point additionally contributes to the scalar terrain-gap residual
\begin{equation}
    r^{\mathrm{terr}}_{k,i}
    =
    z_{c_i,k}
    -
    f_{\boldsymbol{\lambda}}(x_{c_i,k},y_{c_i,k})
    \in\mathbb{R},
    \quad
    i\in\{1,2,3,4\}.
    \label{eqn:zci}
\end{equation}
The four loop-closure residual vectors and four terrain-gap residuals provide 16 scalar residual components for the 15 unknowns in $\boldsymbol{\xi}_k$. We stack them in the explicit order
\begin{equation}
\begin{aligned}
    \mathbf{g}
    (
        \mathbf{x}_k,
        \boldsymbol{\xi}_k,
        \boldsymbol{\lambda}
    )
    =
    \begin{bmatrix}
        \mathbf{r}^{\mathrm{kin}}_{k,1} \\
        \vdots \\
        \mathbf{r}^{\mathrm{kin}}_{k,4} \\
        r^{\mathrm{terr}}_{k,1} \\
        \vdots \\
        r^{\mathrm{terr}}_{k,4}
    \end{bmatrix}
    \in\mathbb{R}^{16}.
\end{aligned}
    \label{eqn:pose_residual_stack}
\end{equation}
All residual components are lengths measured in meters, and the implementation assigns them unit component weights. Because the system is overdetermined, an exact root may not exist. We therefore compute the kinematic best-fit pose/contact-point configuration as the nonlinear least-squares solution
\begin{equation}
    \boldsymbol{\xi}_k^*
    =
    \arg\min_{\boldsymbol{\xi}_k}
    \frac{1}{2}
    \left\|
        \mathbf{g}
        (
            \mathbf{x}_k,
            \boldsymbol{\xi}_k,
            \boldsymbol{\lambda}^*
        )
    \right\|_2^2.
    \label{eqn:pose_nls}
\end{equation}
This solution minimizes the combined loop-closure and terrain-gap discrepancies but does not require either residual group to vanish. Consequently, the optimized $\mathbf{p}_{oc,i}^{*}$ are best-fit contact-point estimates rather than guaranteed load-bearing contacts. The model does not impose unilateral contact, suspension compliance, static force balance, or contact dynamics.

\begin{figure}[t]
  \centering
  \includegraphics[width=0.8\linewidth]{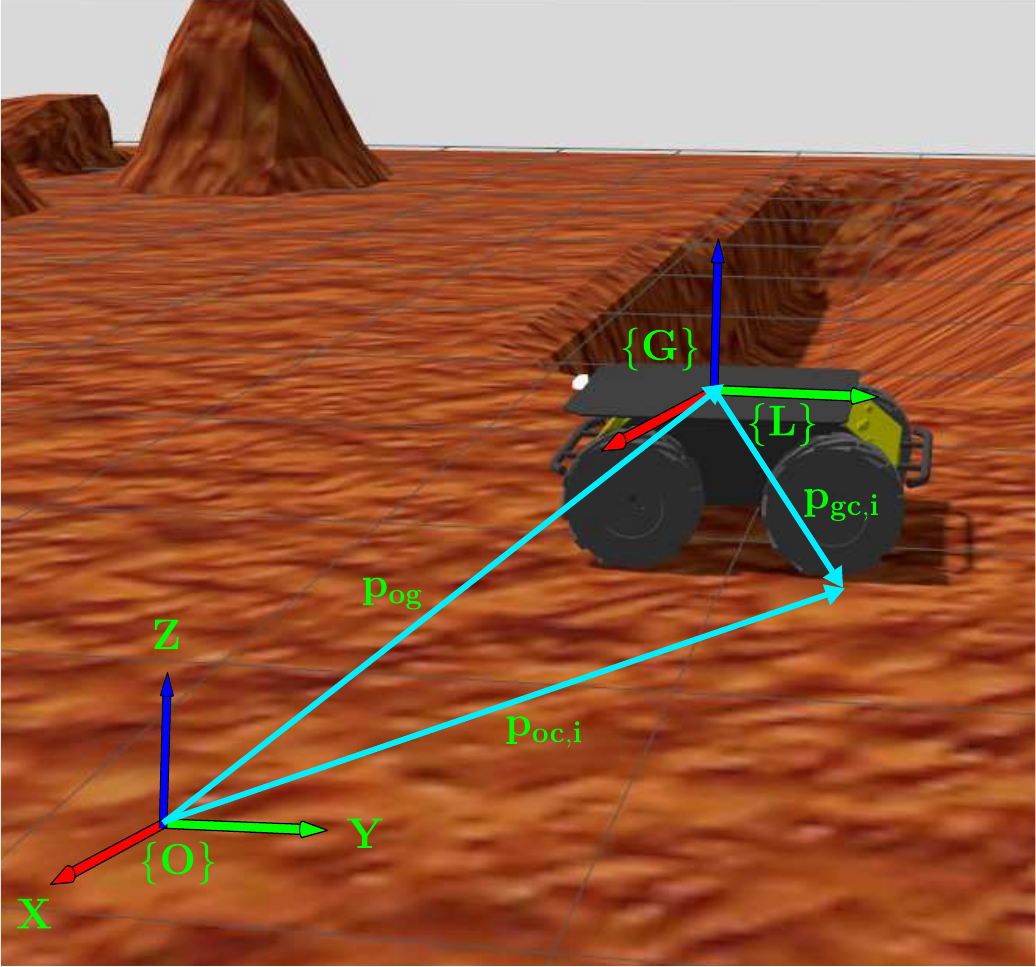}
  \caption{Kinematic model of the wheeled robot, illustrating the global, body-fixed, and translating reference frames used for loop closure.}
  \label{fig:husky}
\end{figure}

This defines the implicit mapping
\begin{equation}
    \boldsymbol{\xi}_k^*
    =
    f_p(\mathbf{x}_k,\boldsymbol{\lambda}^*).
    \label{eqn:x_pose}
\end{equation}

\subsubsection{Implicit Differentiation and Uncertainty Propagation}

\noindent Since $\boldsymbol{\xi}_k^*$ is the solution of an optimization problem, we use implicit differentiation of the optimality conditions to compute its sensitivity with respect to the terrain parameters. Define
\begin{equation}
    E(\boldsymbol{\xi}_k,\boldsymbol{\lambda})
    =
    \frac{1}{2}
    \mathbf{g}
    (
        \mathbf{x}_k,
        \boldsymbol{\xi}_k,
        \boldsymbol{\lambda}
    )^{\mathsf{T}}
    \mathbf{g}
    (
        \mathbf{x}_k,
        \boldsymbol{\xi}_k,
        \boldsymbol{\lambda}
    ).
\end{equation}
At a local optimum $\boldsymbol{\xi}_k^*$, the first-order condition is
\begin{equation}
    \mathbf{F}
    (
        \boldsymbol{\xi}_k^*,
        \boldsymbol{\lambda}^*
    )
    :=
    \nabla_{\boldsymbol{\xi}_k}
    E
    (
        \boldsymbol{\xi}_k^*,
        \boldsymbol{\lambda}^*
    )
    =
    \mathbf{J}_{\boldsymbol{\xi}}^{\mathsf{T}}
    \mathbf{g}
    =
    \mathbf{0},
    \label{eqn:optimality}
\end{equation}
where
\begin{equation}
    \mathbf{J}_{\boldsymbol{\xi}}
    =
    \frac{\partial \mathbf{g}}
    {\partial \boldsymbol{\xi}_k}
    \in
    \mathbb{R}^{16\times 15},
    \qquad
    \mathbf{J}_{\boldsymbol{\lambda}}
    =
    \frac{\partial \mathbf{g}}
    {\partial \boldsymbol{\lambda}}
    \in
    \mathbb{R}^{16\times 2N}.
\end{equation}
Both Jacobians are evaluated at
$(\mathbf{x}_k,\boldsymbol{\xi}_k^*,\boldsymbol{\lambda}^*)$.

Applying the implicit function theorem to \eqref{eqn:optimality} gives
\begin{equation}
    \frac{\mathrm{d}\boldsymbol{\xi}_k^*}
    {\mathrm{d}\boldsymbol{\lambda}}
    =
    -
    \left(
        \frac{\partial \mathbf{F}}
        {\partial \boldsymbol{\xi}_k}
    \right)^{-1}
    \frac{\partial \mathbf{F}}
    {\partial \boldsymbol{\lambda}}.
    \label{eqn:exact_ift}
\end{equation}
The exact derivatives are
\begin{equation}
\begin{aligned}
    \frac{\partial \mathbf{F}}
    {\partial \boldsymbol{\xi}_k}
    &=
    \mathbf{J}_{\boldsymbol{\xi}}^{\mathsf{T}}
    \mathbf{J}_{\boldsymbol{\xi}}
    +
    \sum_{s=1}^{16}
    g_s
    \nabla_{\boldsymbol{\xi}_k\boldsymbol{\xi}_k}^{2} g_s,
    \\
    \frac{\partial \mathbf{F}}
    {\partial \boldsymbol{\lambda}}
    &=
    \mathbf{J}_{\boldsymbol{\xi}}^{\mathsf{T}}
    \mathbf{J}_{\boldsymbol{\lambda}}
    +
    \sum_{s=1}^{16}
    g_s
    \nabla_{\boldsymbol{\xi}_k\boldsymbol{\lambda}}^{2} g_s.
\end{aligned}
\label{eqn:exact_ift_terms}
\end{equation}
Provided the stationary point is locally isolated and $\partial\mathbf{F}/\partial\boldsymbol{\xi}_k$ is nonsingular, \eqref{eqn:exact_ift} defines the local sensitivity of the converged NLS solution. In implementation, JAXopt implicit differentiation evaluates this derivative directly from the optimality condition $\mathbf{F}=\mathbf{0}$, including the residual-weighted terms in \eqref{eqn:exact_ift_terms}, up to the nonlinear- and linear-solver tolerances. We denote this sensitivity by
\begin{equation}
    \mathbf{J}_{\boldsymbol{\lambda},k}^{\boldsymbol{\xi}}
    :=
    \frac{\mathrm{d}\boldsymbol{\xi}_k^*}
    {\mathrm{d}\boldsymbol{\lambda}}.
    \label{eqn:implicit_jacobian}
\end{equation}

Using first-order uncertainty propagation, the inverse-Hessian terrain-parameter estimate gives the following approximate marginal uncertainty matrix for the best-fit pose/contact-point solution:
\begin{equation}
    \boldsymbol{\Sigma}_{\xi_k}
    \approx
    \mathbf{J}_{\boldsymbol{\lambda},k}^{\boldsymbol{\xi}}
    \boldsymbol{\Sigma}_{\lambda}
    \left(
        \mathbf{J}_{\boldsymbol{\lambda},k}^{\boldsymbol{\xi}}
    \right)^{\mathsf{T}}.
    \label{eqn:pose_covariance}
\end{equation}
The diagonal entries of $\boldsymbol{\Sigma}_{\xi_k}$ estimate the marginal uncertainty of the predicted elevation, pitch, roll, and best-fit contact-point estimates. This propagation characterizes the variability of the NLS best-fit output.

Because the same terrain parameters affect every step of a trajectory, the first-order uncertainties at different time steps are coupled. For two time steps $k$ and $\ell$, the corresponding cross-uncertainty estimate is
\begin{equation}
    \boldsymbol{\Sigma}_{\xi_k,\xi_\ell}
    \approx
    \mathbf{J}_{\boldsymbol{\lambda},k}^{\boldsymbol{\xi}}
    \boldsymbol{\Sigma}_{\lambda}
    \left(
        \mathbf{J}_{\boldsymbol{\lambda},\ell}^{\boldsymbol{\xi}}
    \right)^{\mathsf{T}}.
    \label{eqn:cross_time_covariance}
\end{equation}
The planner uses the per-step marginal estimate \eqref{eqn:pose_covariance} for efficient cost evaluation, while \eqref{eqn:cross_time_covariance} describes the corresponding first-order trajectory-level coupling.

\subsection{Trajectory Optimization on Uneven Terrain}
\label{sec:traj_opt}

\noindent We represent the planar vehicle motion using the unicycle model
\begin{equation}
    \dot{x}_k
    =
    v_k\cos\alpha_k,
    \qquad
    \dot{y}_k
    =
    v_k\sin\alpha_k,
    \qquad
    \dot{\alpha}_k
    =
    \Omega_k,
\end{equation}
where $v_k$ and $\Omega_k$ are the linear and angular velocities. Away from zero translational speed, the unicycle is differentially flat with the two-dimensional position as its flat output. Accordingly, the trajectory optimization variables are
\begin{equation}
    \mathbf{p}_k
    =
    \begin{bmatrix}
        x_k & y_k
    \end{bmatrix}^{\mathsf{T}}.
\end{equation}
The measured initial point $\mathbf{p}_0$ is indexed by $k=0$. The planning horizon contains $H$ equal-duration future intervals and $H$ corresponding future samples indexed by $k=1,\ldots,H$, so the complete path is $\{\mathbf{p}_k\}_{k=0}^{H}$ and its duration is $T=H\Delta t$. At each indexed point, define
\begin{equation}
    v_k
    =
    \sqrt{\dot{x}_k^2+\dot{y}_k^2}.
\end{equation}
The initial yaw $\alpha_{\mathrm{odom}}$ is obtained from odometry. At subsequent points $k=1,\ldots,H$, yaw is reconstructed from the path tangent whenever the vehicle is moving and is held at its last valid value when the speed falls below a small threshold $\epsilon_v$:
\begin{align}
    \alpha_0
    &=
    \alpha_{\mathrm{odom}},
    \\
    \alpha_k
    &=
    \begin{cases}
        \operatorname{atan2}(\dot{y}_k,\dot{x}_k),
        & v_k>\epsilon_v,\\
        \alpha_{k-1},
        & v_k\leq\epsilon_v,
    \end{cases}
    \label{eqn:yaw_from_flat_output}
\end{align}
At samples with $v_k\leq\epsilon_v$, including a zero-speed endpoint, the held yaw is supplied to the pose/contact model. The nonnegative speed and tangent-based yaw reconstruction define a forward-motion planning convention. During execution, the downstream path-tracking controller can perform any stationary reorientation required between $\alpha_{\mathrm{odom}}$ and the first moving path tangent. Prescribed endpoint velocities determine the corresponding boundary directions, while headings along the moving portion of the trajectory follow \eqref{eqn:yaw_from_flat_output}.

The trajectory has prescribed endpoint positions and planar velocities. At the trajectory level, the optimization problem is
\begin{equation}
\begin{aligned}
    \min_{x_{1:H},\,y_{1:H}}
    \quad & c_{cem}
    \\
    \mathrm{s.t.}
    \quad &
    x_0=x_{\mathrm{init}},
    \qquad
    y_0=y_{\mathrm{init}},
    \\
    &
    x_H=x_f,
    \qquad
    y_H=y_f,
    \\
    &
    \dot{x}_0=\dot{x}_{\mathrm{init}},
    \qquad
    \dot{y}_0=\dot{y}_{\mathrm{init}},
    \\
    &
    \dot{x}_H=\dot{x}_f,
    \qquad
    \dot{y}_H=\dot{y}_f.
\end{aligned}
\label{eqn:traj_opt}
\end{equation}
Here, $(\dot{x}_{\mathrm{init}},\dot{y}_{\mathrm{init}})$ and $(\dot{x}_f,\dot{y}_f)$ are the prescribed initial and terminal planar velocities, respectively; stationary endpoints are included as a special case.
All trajectory costs are accumulated over the $H$ future samples $k=1,\ldots,H$. The measured state at $k=0$ provides the fixed initial boundary condition, so the CEM array contains only future samples. The implementation evaluates the costs as unnormalized discrete sums using $H=100$ and $\Delta t=0.2\,\mathrm{s}$; the reported objective weights correspond to this discretization. The total objective $c_{cem}$ combines the nominal geometric term developed below with the epistemic-uncertainty penalty derived in Section~\ref{sec:uncertainty_costs}. Curvature and acceleration are computed from the planar trajectory, pose cost from the NLS-predicted roll and pitch, and surface-normal cost from the predicted wheel contacts and fitted terrain gradients. Section~\ref{sec:cem_solution_process} introduces the polynomial parameterization and numerical solution after defining the complete objective.

\subsubsection{Curvature Cost}

The path curvature is
\begin{equation}
    \kappa_k
    =
    \frac{
        \dot{x}_k\ddot{y}_k
        -
        \dot{y}_k\ddot{x}_k
    }{
        (\dot{x}_k^2+\dot{y}_k^2+\epsilon)^{3/2}
    },
    \label{eqn:curvature}
\end{equation}
where $\epsilon>0$ regularizes the curvature calculation near zero speed. Away from rest, where its influence is negligible, the unicycle yaw rate satisfies $\Omega_k=v_k\kappa_k$. The curvature cost is
\begin{equation}
    c_{\kappa}
    =
    \sum_{k=1}^{H}
    \kappa_k^2.
    \label{eqn:curvature_cost}
\end{equation}
The quadratic term $c_{\kappa}$ provides a soft penalty on high-curvature paths.

\subsubsection{Acceleration Cost}

To discourage rapid changes in translational velocity and regularize the sampled trajectories, we penalize planar acceleration:
\begin{equation}
    c_a
    =
    \sum_{k=1}^{H}
    \left(
        \ddot{x}_k^2
        +
        \ddot{y}_k^2
    \right).
    \label{eqn:smoothness_cost}
\end{equation}

\subsubsection{Surface-Normal Cost}

For each NLS best-fit wheel contact-point estimate
$\mathbf{p}_{oc,i,k}=[x_{c_i,k},y_{c_i,k},z_{c_i,k}]^{\mathsf{T}}$,
the local terrain normal is computed from the terrain gradients:
\begin{equation}
    \mathbf{n}_{s,i,k}
    =
    \begin{bmatrix}
        \partial_x f_{\boldsymbol{\lambda}^*}(x_{c_i,k},y_{c_i,k}) \\
        \partial_y f_{\boldsymbol{\lambda}^*}(x_{c_i,k},y_{c_i,k}) \\
        -1
    \end{bmatrix},
    \qquad
    \hat{\mathbf{n}}_{s,i,k}
    =
    \frac{
        \mathbf{n}_{s,i,k}
    }{
        \|\mathbf{n}_{s,i,k}\|_2
    }.
    \label{eqn:terrain_normal}
\end{equation}
Let the flat-ground reference normal be
\begin{equation}
    \mathbf{n}_{\mathrm{ref}}
    =
    \begin{bmatrix}
        0 & 0 & -1
    \end{bmatrix}^{\mathsf{T}}.
\end{equation}
The nominal surface-normal deviation at wheel $i$ and time $k$ is
\begin{equation}
    \ell_{n,k,i}
    =
    \left\|
        \hat{\mathbf{n}}_{s,i,k}
        -
        \mathbf{n}_{\mathrm{ref}}
    \right\|_2^2.
    \label{eqn:normal_stage_cost}
\end{equation}
The nominal surface-normal cost is therefore
\begin{equation}
    c_n
    =
    \sum_{k=1}^{H}
    \sum_{i=1}^{4}
    \ell_{n,k,i}.
    \label{eqn:normal_cost}
\end{equation}

\subsubsection{Pose Cost}

The pose cost penalizes large roll and pitch angles:
\begin{equation}
    c_p
    =
    \sum_{k=1}^{H}
    \left(
        \beta_k^2
        +
        \gamma_k^2
    \right).
    \label{eqn:pose_cost}
\end{equation}

Collecting the four costs gives the nominal geometric objective
\begin{equation}
    c_{nom}
    =
    w_{\kappa}c_{\kappa}
    +
    w_a c_a
    +
    w_n c_n
    +
    w_p c_p.
    \label{eqn:nominal_cem_cost}
\end{equation}
This term evaluates the terrain-dependent quantities at the nominal terrain parameters $\boldsymbol{\lambda}^*$ and forms the deterministic baseline of the planner.

\subsection{Uncertainty-Aware Costs}
\label{sec:uncertainty_costs}

\noindent Terrain uncertainty affects both the predicted vehicle pose and the surface normals used in the trajectory cost. We retain the nominal geometric objective \eqref{eqn:nominal_cem_cost} and augment it with a separate penalty that quantifies these two effects of missing terrain information.

\subsubsection{Uncertainty-Aware Pose Cost}

Let the marginal distribution of the best-fit pose/contact-point state at time $k$ be approximated as
\begin{equation}
    \boldsymbol{\xi}_k
    \sim
    \mathcal{N}
    (
        \boldsymbol{\mu}_{\xi_k},
        \boldsymbol{\Sigma}_{\xi_k}
    ),
\end{equation}
where
\begin{equation}
    \boldsymbol{\mu}_{\xi_k}
    =
    \boldsymbol{\xi}_k^*.
\end{equation}
Let $\mu_{\beta,k}$ and $\mu_{\gamma,k}$ denote the pitch and roll components of $\boldsymbol{\mu}_{\xi_k}$, respectively, and let $\Sigma_{\beta\beta,k}$ and $\Sigma_{\gamma\gamma,k}$ denote the corresponding diagonal entries of $\boldsymbol{\Sigma}_{\xi_k}$. Since the pose stage cost is quadratic, its expectation is available in closed form:
\begin{equation}
    \mathbb{E}
    [
        \beta_k^2+\gamma_k^2
    ]
    =
    \mu_{\beta,k}^2
    +
    \mu_{\gamma,k}^2
    +
    \Sigma_{\beta\beta,k}
    +
    \Sigma_{\gamma\gamma,k}.
    \label{eqn:expected_pose_cost}
\end{equation}
Because the nominal state satisfies $\boldsymbol{\mu}_{\xi_k}=\boldsymbol{\xi}_k^*$, the first two terms in \eqref{eqn:expected_pose_cost} are already included in the nominal pose cost $c_p$. We therefore isolate the additional pose-uncertainty term as
\begin{equation}
    u_p
    =
    \sum_{k=1}^{H}
    \left(
        \Sigma_{\beta\beta,k}
        +
        \Sigma_{\gamma\gamma,k}
    \right).
    \label{eqn:unc_pose_cost}
\end{equation}
Thus, $c_p+u_p$ is the expected quadratic pose cost under the first-order Gaussian approximation.

\subsubsection{Uncertainty-Aware Surface-Normal Cost}

The surface-normal cost depends on the uncertain terrain in two ways. First, the contact location depends on $\boldsymbol{\lambda}$ through the implicit best-fit pose/contact-point solution $\boldsymbol{\xi}_k^*=f_p(\mathbf{x}_k,\boldsymbol{\lambda})$. Second, the terrain normal depends directly on the terrain parameters through the gradients of $f_{\boldsymbol{\lambda}}$.

To account for both effects without linearizing the squared cost at its minimum, define the normal-deviation vector
\begin{equation}
    \mathbf{q}_{n,k,i}
    =
    \widehat{\mathbf{n}}_{s,i,k}
    (
        \boldsymbol{\xi}_k,
        \boldsymbol{\lambda}
    )
    -
    \mathbf{n}_{\mathrm{ref}},
    \qquad
    \mathbf{q}_{n,k,i}^{*}
    =
    \mathbf{q}_{n,k,i}
    (
        \boldsymbol{\xi}_k^{*},
        \boldsymbol{\lambda}^{*}
    ).
\end{equation}
Its total derivative with respect to the terrain parameters is
\begin{equation}
    \mathbf{J}_{q,k,i}
    :=
    \left.
    \frac{\mathrm{d}\mathbf{q}_{n,k,i}}
    {\mathrm{d}\boldsymbol{\lambda}}
    \right|_{
        (\boldsymbol{\xi}_k^{*},\boldsymbol{\lambda}^{*})
    }
    =
    \frac{\partial \mathbf{q}_{n,k,i}}
    {\partial \boldsymbol{\lambda}}
    +
    \frac{\partial \mathbf{q}_{n,k,i}}
    {\partial \boldsymbol{\xi}_k}
    \mathbf{J}_{\boldsymbol{\lambda},k}^{\boldsymbol{\xi}}.
    \label{eqn:normal_total_derivative}
\end{equation}
All partial derivatives in \eqref{eqn:normal_total_derivative} are evaluated at $(\boldsymbol{\xi}_k^{*},\boldsymbol{\lambda}^{*})$. The first term captures the direct dependence of the terrain normal on $\boldsymbol{\lambda}$, while the second term captures the indirect dependence through the contact-point solution.

Using first-order error propagation, the covariance of the normal-deviation vector is approximated by
\begin{equation}
    \boldsymbol{\Sigma}_{q,k,i}
    =
    \mathbf{J}_{q,k,i}
    \boldsymbol{\Sigma}_{\lambda}
    \mathbf{J}_{q,k,i}^{\mathsf{T}}.
    \label{eqn:normal_deviation_covariance}
\end{equation}
Since $\ell_{n,k,i}=\|\mathbf{q}_{n,k,i}\|_2^2$, the expected squared normal deviation under this linearized Gaussian approximation is
\begin{equation}
    \mathbb{E}
    [
        \ell_{n,k,i}
    ]
    \approx
    \left\|
        \mathbf{q}_{n,k,i}^{*}
    \right\|_2^2
    {}+
    \operatorname{tr}
    \left(
        \boldsymbol{\Sigma}_{q,k,i}
    \right).
    \label{eqn:expected_normal_cost}
\end{equation}
The first term is already included in the nominal normal cost $c_n$. We therefore define the additional surface-normal uncertainty term as
\begin{equation}
    u_n
    =
    \sum_{k=1}^{H}
    \sum_{i=1}^{4}
    \operatorname{tr}
    \left(
        \boldsymbol{\Sigma}_{q,k,i}
    \right).
    \label{eqn:unc_normal_cost}
\end{equation}
Consequently, $c_n+u_n$ approximates the expected aggregate squared normal-deviation cost. Unlike first-order propagation of the scalar squared cost, this correction remains sensitive when the nominal normal equals $\mathbf{n}_{\mathrm{ref}}$ but its terrain-induced perturbation is uncertain.

The weighted epistemic-uncertainty penalty and final CEM objective are
\begin{equation}
    \begin{aligned}
        c_{unc}
        &=
        w_n\rho_n u_n
        +
        w_p\rho_p u_p,
        \\
        c_{cem}
        &=
        c_{nom}
        +
        c_{unc}.
    \end{aligned}
    \label{eqn:final_cem_cost}
\end{equation}
Here, $\rho_n,\rho_p\geq0$ control the strengths of the normal and pose uncertainty penalties. When $\rho_n=\rho_p=1$, the combinations $c_n+u_n$ and $c_p+u_p$ recover the corresponding expected quadratic costs under the first-order Gaussian approximation; values greater than one add further conservatism. The nominal term penalizes geometrically difficult trajectories, while $c_{unc}$ discourages predictions that depend strongly on weakly observed terrain; setting $c_{unc}=0$ recovers the nominal deterministic baseline.

\subsection{Solution Process: Polynomial CEM}
\label{sec:cem_solution_process}
\label{sec:cem_traj_gen}

The implemented solver uses CEM~\cite{rubinstein1999cross}, as shown in Fig.~\ref{fig:cem_pipeline}. Candidate trajectories are parameterized by degree-10 Bernstein polynomials in the unicycle flat outputs introduced in Section~\ref{sec:traj_opt}. This formulation reduces the search from $H$ future positions per planar coordinate to 11 coefficients per coordinate, while enforcing the endpoint position and velocity conditions directly in coefficient space.

Define the normalized sample coordinate
\begin{equation}
    \tau_k=\frac{k}{H},
    \qquad
    k=0,\ldots,H,
    \qquad
    T=H\Delta t.
    \label{eqn:normalized_horizon}
\end{equation}
For $j=0,\ldots,10$, the Bernstein basis and its vector form are
\begin{equation}
\begin{aligned}
    B_j^{10}(\tau)
    &=
    \binom{10}{j}
    \tau^j
    (1-\tau)^{10-j},
    \\
    \mathbf{w}(\tau)
    &=
    \begin{bmatrix}
        B_0^{10}(\tau) & \cdots & B_{10}^{10}(\tau)
    \end{bmatrix}^{\mathsf{T}}.
\end{aligned}
    \label{eqn:bernstein_basis}
\end{equation}
Thus, $n_c=11$. The basis and physical-time derivative matrices evaluated at the $H$ future samples are defined row-wise by
\begin{equation}
\begin{aligned}
    \mathbf{W}_{k,:}
    &=
    \mathbf{w}(\tau_k)^{\mathsf{T}},
    \\
    \dot{\mathbf{W}}_{k,:}
    &=
    \frac{1}{T}
    \left(
        \left.
        \frac{\mathrm{d}\mathbf{w}}
        {\mathrm{d}\tau}
        \right|_{\tau=\tau_k}
    \right)^{\mathsf{T}},
    \\
    \ddot{\mathbf{W}}_{k,:}
    &=
    \frac{1}{T^2}
    \left(
        \left.
        \frac{\mathrm{d}^2\mathbf{w}}
        {\mathrm{d}\tau^2}
        \right|_{\tau=\tau_k}
    \right)^{\mathsf{T}},
    \qquad
    k=1,\ldots,H.
\end{aligned}
\label{eqn:bernstein_matrices}
\end{equation}
Hence, $\mathbf{W},\dot{\mathbf{W}},\ddot{\mathbf{W}}\in\mathbb{R}^{H\times11}$. For $\mathbf{c}_x,\mathbf{c}_y\in\mathbb{R}^{11}$, the $H$ future positions are
\begin{equation}
    \mathbf{P}_{1:H}
    :=
    \begin{bmatrix}
        \mathbf{p}_1^{\mathsf{T}}\\
        \vdots\\
        \mathbf{p}_H^{\mathsf{T}}
    \end{bmatrix}
    =
    \begin{bmatrix}
        \mathbf{W}\mathbf{c}_x &
        \mathbf{W}\mathbf{c}_y
    \end{bmatrix}
    \in\mathbb{R}^{H\times2}.
    \label{eqn:coefficient_trajectory}
\end{equation}

The endpoint position and velocity conditions are imposed in coefficient space using
\begin{equation}
\begin{aligned}
    \mathbf{A}
    &=
    \begin{bmatrix}
        \mathbf{w}(0)^{\mathsf{T}}\\
        T^{-1}\mathbf{w}'(0)^{\mathsf{T}}\\
        \mathbf{w}(1)^{\mathsf{T}}\\
        T^{-1}\mathbf{w}'(1)^{\mathsf{T}}
    \end{bmatrix},
    \\
    \mathbf{b}_x
    &=
    \begin{bmatrix}
        x_{\mathrm{init}} &
        \dot{x}_{\mathrm{init}} &
        x_f &
        \dot{x}_f
    \end{bmatrix}^{\mathsf{T}},
    \\
    \mathbf{b}_y
    &=
    \begin{bmatrix}
        y_{\mathrm{init}} &
        \dot{y}_{\mathrm{init}} &
        y_f &
        \dot{y}_f
    \end{bmatrix}^{\mathsf{T}},
\end{aligned}
    \label{eqn:coefficient_boundary_constraints}
\end{equation}
where primes denote derivatives with respect to $\tau$. The implemented restriction of \eqref{eqn:traj_opt} is therefore
\begin{equation}
\begin{aligned}
    \min_{\mathbf{c}_x,\mathbf{c}_y\in\mathbb{R}^{11}}
    \quad &
    c_{cem}
    \left(
        \mathbf{W}\mathbf{c}_x,
        \mathbf{W}\mathbf{c}_y
    \right)
    \\
    \mathrm{s.t.}
    \quad &
    \mathbf{A}\mathbf{c}_x=\mathbf{b}_x,
    \qquad
    \mathbf{A}\mathbf{c}_y=\mathbf{b}_y.
\end{aligned}
\label{eqn:coefficient_traj_opt}
\end{equation}
The derivative-dependent costs use $\dot{\mathbf{W}}\mathbf{c}_{x/y}$ and $\ddot{\mathbf{W}}\mathbf{c}_{x/y}$.

CEM maintains a Gaussian distribution over the joint coefficient vector
\begin{equation}
    \mathbf{c}
    =
    \begin{bmatrix}
        \mathbf{c}_x^{\mathsf{T}} &
        \mathbf{c}_y^{\mathsf{T}}
    \end{bmatrix}^{\mathsf{T}}
    \in\mathbb{R}^{22}.
\end{equation}
For each sampled coordinate vector $\widetilde{\mathbf{c}}_q$, $q\in\{x,y\}$, a projection filter solves
\begin{equation}
\begin{aligned}
    \overline{\mathbf{c}}_q
    &=
    \underset{\mathbf{A}\mathbf{c}=\mathbf{b}_q}
    {\operatorname*{arg\,min}}
    \frac{1}{2}
    \left\|
        \mathbf{c}-\widetilde{\mathbf{c}}_q
    \right\|_2^2
    \\
    &=
    \widetilde{\mathbf{c}}_q
    -
    \mathbf{A}^{\mathsf{T}}
    \left(
        \mathbf{A}\mathbf{A}^{\mathsf{T}}
    \right)^{-1}
    \left(
        \mathbf{A}\widetilde{\mathbf{c}}_q
        -
        \mathbf{b}_q
    \right),
\end{aligned}
\label{eqn:coefficient_projection}
\end{equation}
where $\mathbf{A}$ has full row rank. Thus, every evaluated candidate satisfies the boundary conditions.

Before the first iteration, CEM uses the Bernstein coefficients of the boundary-consistent straight-line path as its initial mean. Its covariance is the empirical covariance of Bernstein fits to 100 straight-line paths with smoothness-shaped lateral noise, regularized by $10^{-3}\mathbf{I}$. Before CEM begins, the terrain parameters $\boldsymbol{\lambda}^*$ and their uncertainty $\boldsymbol{\Sigma}_{\lambda}$ are computed once from the current point cloud using \eqref{eqn:terrain_fit} and \eqref{eqn:sigma_lambda}. These quantities remain fixed throughout the CEM search. At each iteration, CEM draws $n_s$ coefficient samples and projects them onto the endpoint constraints using \eqref{eqn:coefficient_projection}. Each projected trajectory is evaluated using $c_{cem}$ from \eqref{eqn:final_cem_cost}. This evaluation solves the NLS pose/contact problem \eqref{eqn:pose_nls} at every waypoint and computes the associated uncertainty terms using \eqref{eqn:implicit_jacobian}--\eqref{eqn:pose_covariance} and \eqref{eqn:normal_total_derivative}--\eqref{eqn:normal_deviation_covariance}.

The $n_e$ lowest-cost candidates update the coefficient distribution at each iteration. After $n_{cem}$ iterations, the planner returns the lowest-cost planar path $\{\mathbf{p}_k\}_{k=0}^{H}$. Its terrain-surface representation is $\{(x_k,y_k,f_{\boldsymbol{\lambda}^*}(x_k,y_k))\}_{k=0}^{H}$, and the corresponding predicted body trajectory used for cost evaluation is $\{(x_k,y_k,z_k^*,\alpha_k,\beta_k^*,\gamma_k^*)\}_{k=0}^{H}$.

\section{Flow Matching Warm Start for Terrain Fitting}\label{flow_acc}
To reduce terrain-fitting latency, we train a conditional Flow Matching model offline that generates a single initialization for terrain-parameter refinement~\cite{tong2023conditional,lipman2024flow}.

Let $\boldsymbol{\zeta}_1^{(\lambda)}\in\mathbb{R}^{d_{\lambda}}$ denote the training target with condition $\mathbf{c}_{\lambda}$, and draw a source sample $\boldsymbol{\zeta}_0^{(\lambda)}\sim\mathcal{N}(\mathbf{0},\mathbf{I}_{d_{\lambda}})$. Given $t\sim\mathcal{U}[0,1]$, we use the linear conditional probability path
\begin{equation}
    \boldsymbol{\zeta}_t^{(\lambda)}
    =
    (1-t)\boldsymbol{\zeta}_0^{(\lambda)}
    +
    t\boldsymbol{\zeta}_1^{(\lambda)},
    \qquad
    \mathbf{u}_t^{(\lambda)}
    =
    \boldsymbol{\zeta}_1^{(\lambda)}-\boldsymbol{\zeta}_0^{(\lambda)} ,
    \label{eqn:fm_path}
\end{equation}
where $\mathbf{u}_t^{(\lambda)}=\mathrm{d}\boldsymbol{\zeta}_t^{(\lambda)}/\mathrm{d}t$ is the corresponding conditional target velocity. A velocity field $\mathbf{v}_{\boldsymbol{\theta}_\lambda}$ is trained using
\begin{equation}
    \mathcal{L}_{\mathrm{FM}}^{(\lambda)}
    =
    \mathbb{E}
    \left[
    \left\|
    \mathbf{v}_{\boldsymbol{\theta}_\lambda}
    \left(\boldsymbol{\zeta}_t^{(\lambda)},t,\mathbf{c}_\lambda\right)
    -
    \mathbf{u}_t^{(\lambda)}
    \right\|_2^2
    \right].
    \label{eqn:fm_loss}
\end{equation}
At inference, the learned field is integrated from the Gaussian source to the output:
\begin{equation}
    \frac{\mathrm{d}\boldsymbol{\zeta}_t^{(\lambda)}}{\mathrm{d}t}
    =
    \mathbf{v}_{\boldsymbol{\theta}_\lambda}
    \left(\boldsymbol{\zeta}_t^{(\lambda)},t,\mathbf{c}_\lambda\right),
    \qquad t\in[0,1].
    \label{eqn:fm_ode}
\end{equation}
At deployment, the learned velocity field transports a Gaussian source state to a terrain-parameter warm start for Levenberg--Marquardt refinement. The following paragraphs describe the conditioning, output space, and deployment procedure for the model.

\begin{figure*}[!t]
    \centering
    \includegraphics[width=0.8\linewidth]{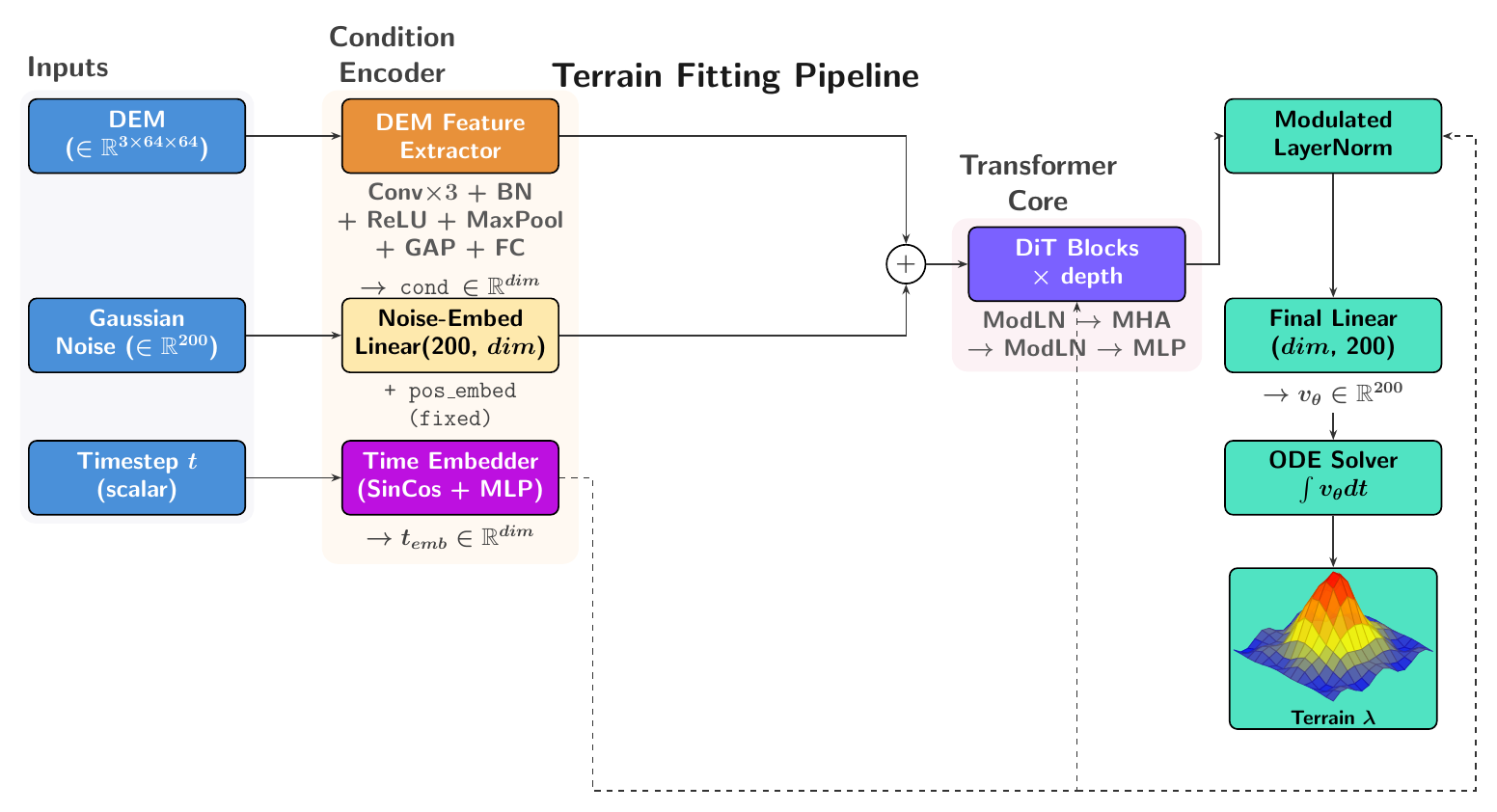}
    \caption{Terrain Parameter Network. A CNN encodes the Digital Elevation Model (DEM) constructed from the current point cloud. Conditioned on this encoding and the flow timestep, the DiT represents a velocity field whose ODE transports a Gaussian initial state to a Fourier-coefficient warm start; Levenberg--Marquardt subsequently refines this output. Abbreviations: Conv -- convolutional layer; BN -- batch normalization; ReLU -- rectified linear unit; MaxPool -- max pooling; GAP -- global average pooling; FC -- fully connected layer; pos embed -- positional embedding; DiT -- Diffusion Transformer; ModLN -- modulated layer normalization; MHA -- multi-head self-attention; MLP -- multi-layer perceptron; SinCos -- sinusoidal timestep encoding; ODE -- ordinary differential equation.}
    \label{fig:terrain_param_network}
\end{figure*}

The Terrain Parameter Network in Fig.~\ref{fig:terrain_param_network} instantiates \eqref{eqn:fm_path}--\eqref{eqn:fm_ode} in the $2N$-dimensional Fourier-coefficient space, so $d_{\lambda}=2N$. Because the Fourier frequencies in Section~\ref{sec:terrain_model} are sampled once and held fixed across all terrain maps, every coefficient coordinate has the same meaning throughout the training set. For each training example, the DEM constructed from a point cloud $\mathcal{D}$ is paired with the coefficient vector obtained by solving \eqref{eqn:terrain_fit} offline; this fitted vector defines the terminal target $\boldsymbol{\zeta}_1^{(\lambda)}$. A CNN feature extractor maps the DEM to the condition $\mathbf{c}_{\lambda}$. The evolving coefficient state $\boldsymbol{\zeta}_t^{(\lambda)}$ and timestep $t$ are then embedded and processed by a Diffusion Transformer (DiT)~\cite{peebles2023scalable}, which parameterizes the conditional velocity field $\mathbf{v}_{\boldsymbol{\theta}_{\lambda}}$.

At deployment, the DEM constructed from the current point cloud provides the condition $\mathbf{c}_{\lambda}$. Starting from one state $\boldsymbol{\zeta}_0^{(\lambda)}\sim\mathcal{N}(\mathbf{0},\mathbf{I}_{2N})$, integration of \eqref{eqn:fm_ode} produces a terrain-parameter warm start. One to two Levenberg--Marquardt iterations refine this prediction against the pointwise residuals in \eqref{eqn:terrain_fit}, yielding $\boldsymbol{\lambda}^*$. The refined parameters define the nominal surface $f_{\boldsymbol{\lambda}^*}$, and the regularized Hessian of the current fitting problem provides $\boldsymbol{\Sigma}_{\lambda}$. These refined model-based quantities are then used for pose/contact prediction and uncertainty-aware trajectory evaluation.

%% file: Results.tex
\section{Validation and Benchmarking}

The objective of this section is to answer the following research questions:

\begin{itemize}
\item \textbf{Q1:} How do route selection and closed-loop outcomes change when propagated-uncertainty scoring is removed, both qualitatively and across the broader simulation campaign? See Sections~\ref{uncert_effect} and~\ref{quant_comp}.
\item \textbf{Q2:} How does the complete framework compare with Bi-level-opt and GP-navigation under the matched benchmark? See Sections~\ref{qual_comp} and~\ref{quant_comp}.
\item \textbf{Q3:} What end-to-end feasibility evidence is provided by the selected hardware executions? See Section~\ref{real_world_demo}.
\item \textbf{Q4:} What fitting-accuracy--latency trade-off is produced by the FM warm start? See Section~\ref{terrain_fm}.
\end{itemize}

\subsection{Implementation Details}
\noindent The proposed algorithms were implemented in Python, leveraging JAX~\cite{jax2018github} and JAXopt~\cite{jaxopt_implicit_diff} for GPU-accelerated computations, alongside Equinox~\cite{kidger2021equinox} for neural network architectures. The polynomial basis matrix $\mathbf{W}$ uses a degree-10 Bernstein basis, giving $n_c=11$ coefficients per planar coordinate. The evaluation pipeline consists of open-loop analyses conducted via standalone Python scripts, alongside closed-loop simulations executed within Gazebo~\cite{koenig2004design} using the Robot Operating System (ROS)~\cite{quigley2009ros}. Synthetic terrains were modeled in Blender~\cite{blender} and converted into compatible formats for both simulation platforms. All simulation evaluations were performed on a workstation equipped with an Intel Core i9 CPU and an NVIDIA RTX 5090 GPU.

Terrain geometry is reconstructed from a single point-cloud frame using LiDAR and depth camera data, with maximum ranges capped at 10\,m and 5\,m, respectively. Before terrain fitting, the point cloud is voxel-filtered. If more than 35,000 filtered points remain, 35,000 distinct points are selected randomly without replacement. The number of Fourier frequencies is fixed at $N=100$ in all reported experiments. Each planned trajectory used $H=100$ future intervals with $\Delta t=0.2\,\text{s}$, giving the horizon duration $T=H\Delta t=20\,\text{s}$. The measured state at $k=0$ was supplied separately, while the optimizer represented the $H$ future samples at $k=1,\ldots,H$. For trajectory optimization, CEM used $n_{\text{cem}} = 5$ iterations, a population size of $n_s = 100$, and an elite set size of $n_e = 25$.

The terrain-parameter network was trained on a combined point-cloud dataset comprising Gazebo simulation captures, scans from the M2UD off-road dataset~\cite{jia2025m2udmultimodelmultiscenariouneventerrain}, and hardware data collected by us with the Clearpath Husky and Jackal~\cite{clearpathrobotics} platforms. Additional dataset, training, and inference details are provided in Table~\ref{tab:fm_reproducibility_details} of the Appendix.

Hardware experiments were conducted on Clearpath Husky and Jackal~\cite{clearpathrobotics} mobile platforms equipped with an Intel RealSense D435~\cite{realsense_d435} depth camera. Point clouds for the reported hardware demonstrations were collected during these physical runs and processed using the same terrain-reconstruction pipeline as in simulation. Onboard state estimation was achieved via an IMU-based inertial navigation system. Trajectories were executed using a $\pi$-MPPI controller~\cite{pi-mppi}, which generates linear and angular velocity commands for path tracking. Onboard planning computations were executed on an attached laptop equipped with an NVIDIA RTX 3090 GPU.

For each reported closed-loop simulation and hardware trial, the planner generated one terrain-aware reference path from the initial point-cloud frame. No high-level replanning was performed. The resulting path was tracked in closed loop by the $\pi$-MPPI controller, which generated linear and angular velocity commands.

\subsection{Baseline Methods and Evaluation Metrics}
\noindent We compare our approach against two representative baselines. Both baselines were evaluated using the open-source implementations released by their respective authors. For every matched trial, all planners received the same partial terrain sensor observations and start--goal pair. No wall-clock planning-time budget was imposed on either baseline.

\begin{itemize}
\item \textbf{Bi-level-opt~\cite{manoharan2024bi}:} This method shares the NLS-based wheel--terrain pose/contact predictor and implicit-differentiation structure used by our nominal model. Its projected-gradient outer optimizer minimizes deterministic kinematic and tip-over stability costs. However, it treats the fitted terrain as deterministic and does not account for uncertainty caused by missing observation support.
\item \textbf{GP-navigation~\cite{leininger2024gaussian}:} This method constructs sparse-GP elevation and predictive-variance maps and combines slope, flatness, and step-height features into a traversability map. Cells whose predictive variance exceeds a fixed threshold are masked as non-traversable, after which RRT* performs footprint-aware planning over the remaining map. Thus, terrain uncertainty affects planning through a thresholded map-level exclusion rather than through continuous propagation into vehicle-configuration-dependent pose and surface-normal costs.
\end{itemize}

The evaluation includes both qualitative and quantitative assessments. Qualitatively, we evaluate whether planned trajectories avoid occluded or sparsely observed regions, maintain fidelity to the ground-truth geometry, and favor routes with lower predicted vehicle--terrain interaction costs under model uncertainty. Quantitatively, performance in closed-loop simulations is measured using the following metrics:

\begin{itemize}
\item \textbf{Failure Rate:} A run is classified as a failure if the vehicle does not reach the goal within the allotted time, becomes immobilized, or exceeds a roll or pitch magnitude of $40^\circ$.
\item \textbf{RMS and Max Pitch/Roll:} For successful runs, we report the root mean square (RMS) and peak pitch and roll angles to quantify attitude variation along the executed trajectory.
\end{itemize}

Together, these metrics evaluate task completion and attitude variation under partial terrain observability.

\subsection{Effect of Propagated-Uncertainty Scoring} \label{uncert_effect}
\begin{figure}
    \centering
    \includegraphics[width=\columnwidth]{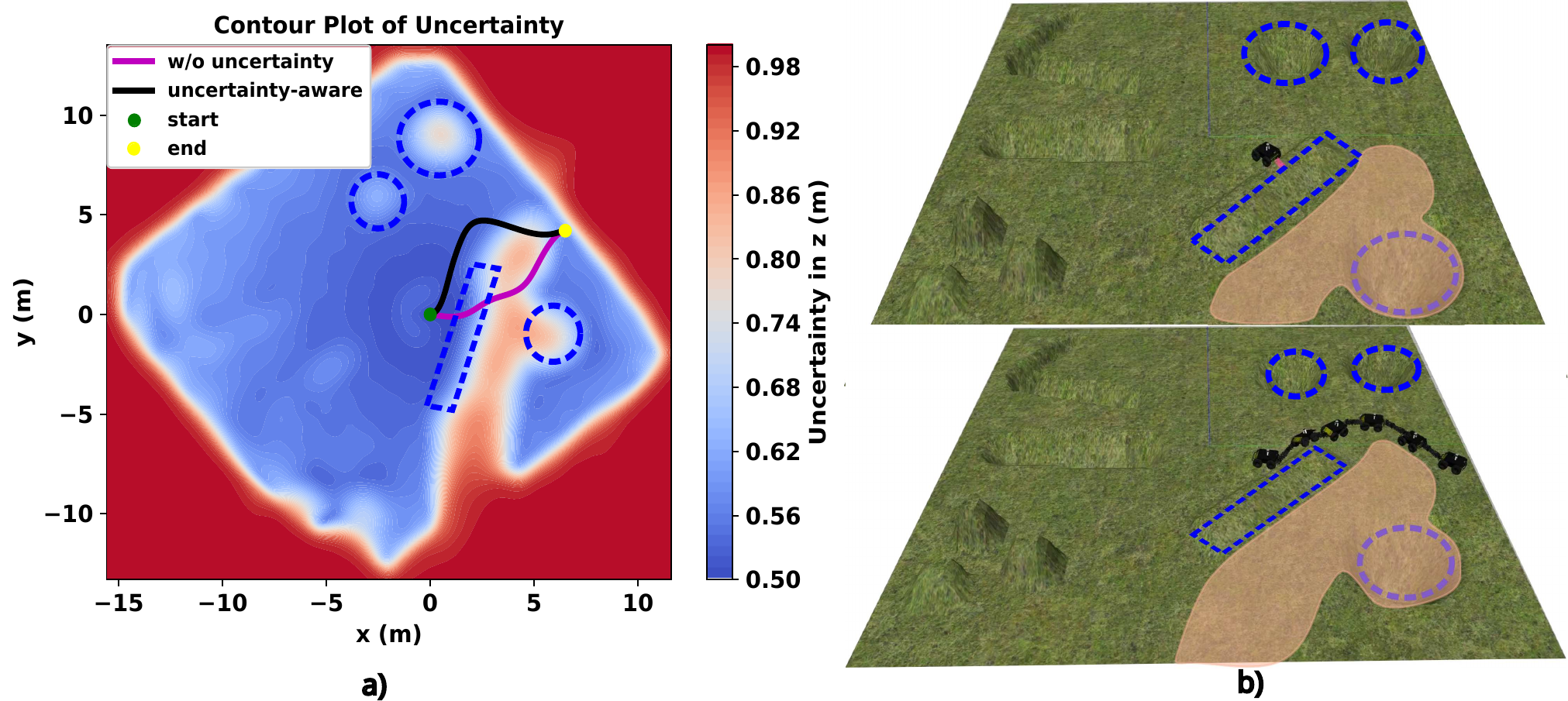}
    \caption{Comparison with the propagated-uncertainty-scoring ablation. Both planners use the same CEM initialization. a) Planned trajectories overlaid on the terrain-uncertainty map. The $c_{unc}=0$ planner (magenta) traverses the high-uncertainty region formed by occlusion behind the bump (blue rectangle), whereas the uncertainty-aware planner (black) routes around it. Blue circles identify the craters. b) Closed-loop Gazebo executions: the $c_{unc}=0$ planner becomes immobilized, while the uncertainty-aware planner reaches the goal.}
    \label{fig:traj1_flow_vs_base_unc_comparison}
\end{figure}

\begin{figure}
    \centering
    \includegraphics[width=\columnwidth]{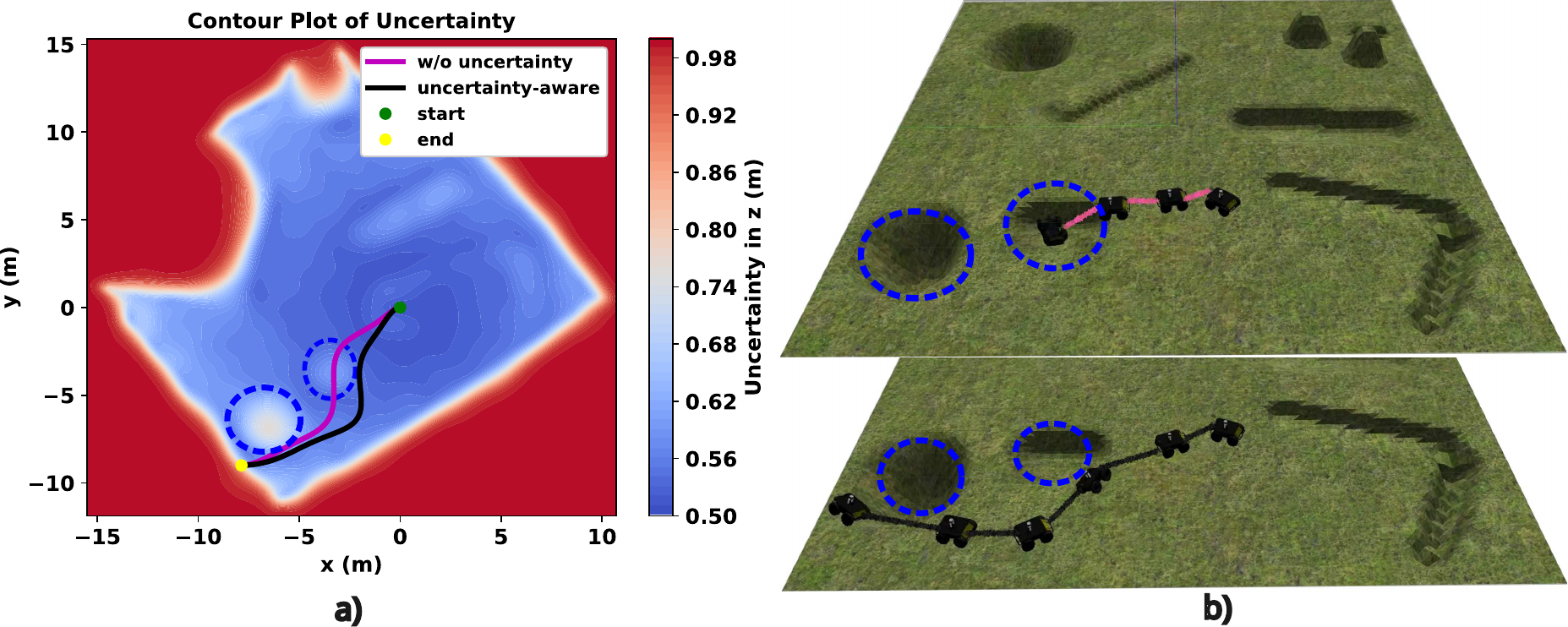}
    \caption{Second comparison with the propagated-uncertainty-scoring ablation. Both planners use the same CEM initialization. a) The $c_{unc}=0$ trajectory (magenta) crosses a high-uncertainty, crater-adjacent region, while the uncertainty-aware trajectory (black) avoids the marked regions. b) In closed-loop Gazebo execution, the $c_{unc}=0$ planner falls into a crater, whereas the uncertainty-aware planner reaches the goal.}
    \label{fig:traj3_flow_vs_base_unc_comparison}
\end{figure}

\noindent We isolate the effect of propagated-uncertainty scoring by comparing two planners that use the same CEM initialization. The uncertainty-aware planner uses $c_{cem}=c_{nom}+c_{unc}$, whereas the ablated planner sets $c_{unc}=0$ and therefore uses only $c_{nom}$. Both planners otherwise use the same partial terrain observations, fitted terrain representation, NLS pose/contact predictor, nominal objective, CEM configuration, and start--goal pair. Differences between their trajectories can therefore be attributed to the inclusion of $c_{unc}$. In Figs.~\ref{fig:traj1_flow_vs_base_unc_comparison} and \ref{fig:traj3_flow_vs_base_unc_comparison}, black denotes the uncertainty-aware planner and magenta denotes the $c_{unc}=0$ ablation.

Fig.~\ref{fig:traj1_flow_vs_base_unc_comparison} shows a case in which the bump marked by the blue rectangle occludes terrain along the more direct route. The $c_{unc}=0$ planner passes through this weakly supported region, whereas the uncertainty-aware planner takes a lower-uncertainty route around it. During closed-loop execution, the $c_{unc}=0$ planner becomes immobilized, whereas the uncertainty-aware planner reaches the goal. Because both planners use the same CEM initialization, the trajectory difference isolates the effect of including propagated-uncertainty scoring in the objective.

The second example, shown in Fig.~\ref{fig:traj3_flow_vs_base_unc_comparison}, exhibits a similar difference in another terrain configuration. The $c_{unc}=0$ trajectory cuts through a high-uncertainty region adjacent to a crater and subsequently fails during execution. The uncertainty-aware planner instead follows a wider route through better-supported terrain and reaches the goal.

\begin{summarybox}
Together, these examples isolate the effect of propagated-uncertainty scoring under a fixed CEM initialization. Including $c_{unc}$ changes the objective so that routes through weakly supported terrain can be rejected in favor of better-supported alternatives. Section~\ref{quant_comp} reports the aggregate effect of this same ablation across the broader simulation campaign.
\end{summarybox}

\subsection{Qualitative Comparison with Baselines} \label{qual_comp}

\begin{figure*}[!t]
    \centering
    \includegraphics[width=\linewidth]{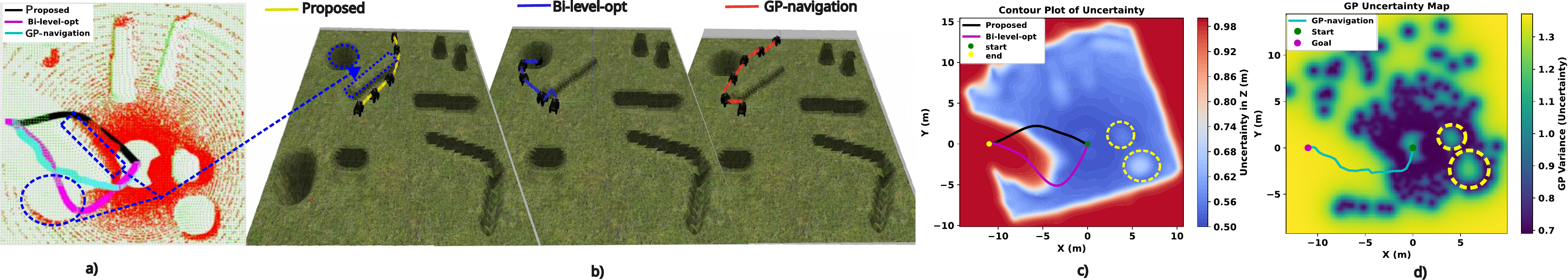}
    \caption{Qualitative comparison in Scene 1. (a)~Planned trajectories overlaid on ground-truth terrain points (green) and partial LiDAR observations (red), with a dashed blue circle and rectangle marking the crater and occluding bump. (b)~Closed-loop Gazebo executions: the proposed method and GP-navigation reach the goal, whereas Bi-level-opt falls into the crater. (c)~Planned trajectories overlaid on the proposed terrain-uncertainty map. (d)~GP predictive-variance map with the GP-navigation path. Dashed yellow circles highlight spatially corresponding high-uncertainty regions in (c) and (d).}
    \label{fig:traj3_baseline_comparison}
\end{figure*}

\begin{figure*}[!t]
    \centering
    \includegraphics[width=\linewidth]{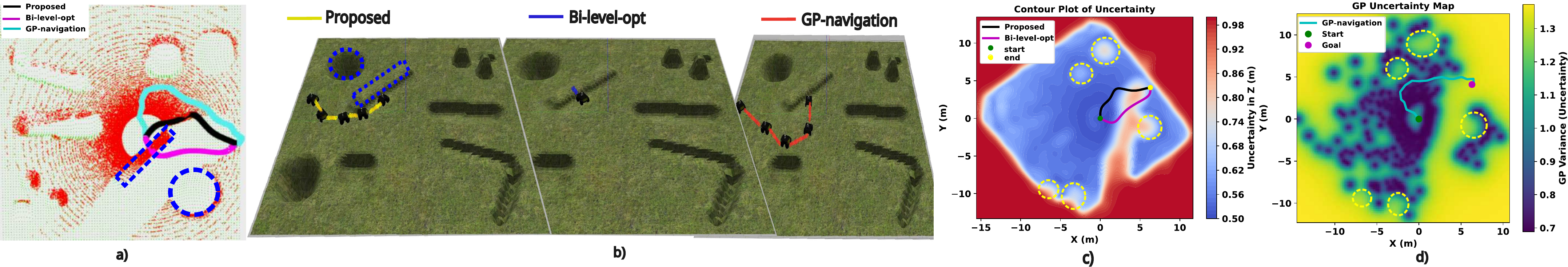}
    \caption{Qualitative comparison in Scene 2. (a)~Planned trajectories overlaid on ground-truth terrain points (green) and partial LiDAR observations (red); dashed annotations mark the bump--crater occlusion. (b)~Closed-loop Gazebo executions: the proposed planner and GP-navigation reach the goal, while Bi-level-opt becomes immobilized on the bumpy terrain. (c)~Planned trajectories overlaid on the proposed terrain-uncertainty map. (d)~GP predictive-variance map and GP-navigation path. Dashed yellow circles highlight corresponding high-uncertainty regions in (c) and (d).}
    \label{fig:traj1_baseline_comparison}
\end{figure*}

\begin{figure*}[!t]
    \centering
    \includegraphics[width=\linewidth]{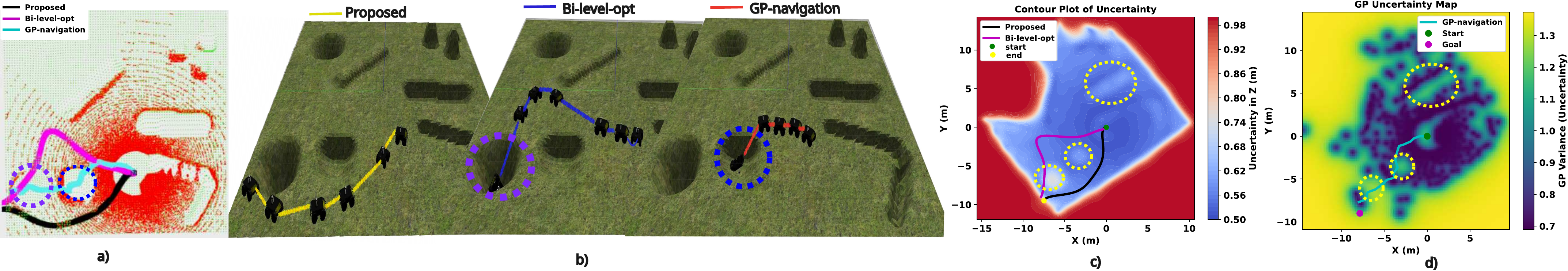}
    \caption{Qualitative comparison in Scene 3. (a)~Planned trajectories overlaid on ground-truth terrain points (green) and partial LiDAR observations (red). (b)~Closed-loop Gazebo executions: the proposed planner avoids both marked hazards and reaches the goal; GP-navigation falls into the crater marked in blue, while Bi-level-opt falls into the crater marked in purple. (c)~Planned trajectories overlaid on the proposed terrain-uncertainty map. (d)~GP predictive-variance map and GP-navigation path. Dashed yellow circles highlight corresponding high-uncertainty regions in (c) and (d).}
    \label{fig:traj_1_1_3_baseline_comparison}
\end{figure*}

\noindent Figures~\ref{fig:traj3_baseline_comparison}--\ref{fig:traj_1_1_3_baseline_comparison} present qualitative comparisons between our framework, Bi-level-opt, and GP-navigation across three matched synthetic scenes. In each scenario, all planners operate on identical partial LiDAR observations and identical start--goal pairs. For each figure:
\begin{itemize}
    \item \textbf{Panel (a)} overlays planned trajectories onto the composite terrain evidence, displaying ground-truth elevation points in green and raw LiDAR returns in red;
    \item \textbf{Panel (b)} illustrates the corresponding closed-loop Gazebo executions;
    \item \textbf{Panel (c)} overlays all planned paths on our coverage-induced terrain-uncertainty map; and
    \item \textbf{Panel (d)} depicts the GP predictive-variance map alongside the GP-navigation trajectory.
\end{itemize}
Across all panels, planned trajectories are color-coded in black (Proposed), magenta (Bi-level-opt), and cyan (GP-navigation), whereas closed-loop executed trajectories are shown in yellow, blue, and red, respectively. Because our regularized inverse-Hessian uncertainty metric and the GP predictive variance have different formulations and numerical scales, their spatial patterns are compared qualitatively rather than by raw magnitude.

Figure~\ref{fig:traj3_baseline_comparison} illustrates Scene~1, where all three planners successfully bypass the central high-elevation obstacle but exhibit distinct behaviors near the occluded bump--crater region. As shown in Panels~(c) and~(d), both our uncertainty estimator and the GP model identify high uncertainty within this weakly observed region. However, while our planner selects a path anchored in well-supported terrain, both Bi-level-opt and GP-navigation enter the uncertain zone. During closed-loop execution (Panel~b), Bi-level-opt falls directly into the unobserved crater, leading to failure. GP-navigation narrowly circumvents the hazard to reach the goal, whereas our proposed approach remains separated from the marked uncertain region throughout. This scene illustrates that, on a route through weakly observed terrain, the execution outcome can differ from the benign outcome suggested by the nominal fit.

Figure~\ref{fig:traj1_baseline_comparison} depicts Scene~2, illustrating a scenario where both uncertainty-aware planners demonstrate similar avoidance behavior. Panels~(c) and~(d) reveal a localized bump--crater occlusion that induces high uncertainty in both models. Both the proposed planner and GP-navigation route around this unobserved zone. In contrast, Bi-level-opt—which plans purely on the deterministic terrain fit—routes directly through the unobserved region, assuming a smooth surface interpolation. During closed-loop execution (Panel~b), Bi-level-opt becomes immobilized on the unmodeled rough terrain, whereas both uncertainty-aware methods complete the traversal. In this trial, the two uncertainty-aware planners reject the weakly supported nominal extrapolation selected by the deterministic baseline.

Scene~3, presented in Fig.~\ref{fig:traj_1_1_3_baseline_comparison}, shows different routes and outcomes under the three planners. As seen in Panels~(c) and~(d), two separate high-uncertainty regions are present. Our proposed method plans around both uncertain zones and reaches the goal. In contrast, GP-navigation traverses the first uncertain zone and falls into an unobserved crater (marked in blue). Bi-level-opt avoids the first region but cuts through the second, subsequently tumbling into another unobserved depression (marked in purple), as shown in Panel~(b). In this scene, the different uncertainty-handling strategies produce distinct routes and outcomes: the proposed planner avoids both uncertain regions, whereas each baseline enters one of them and fails.

\begin{summarybox}
Across the three matched scenes, the proposed planner consistently selected routes with stronger observation support and reached the goal. The baselines entered weakly observed regions in several cases and subsequently became immobilized or fell into unobserved terrain. These route choices illustrate the practical distinction between evaluating a deterministic terrain fit, applying a map-level uncertainty threshold, and using the proposed graded penalty propagated through the predicted pose and individual wheel contacts. Section~\ref{quant_comp} examines whether this qualitative advantage persists across the full matched simulation campaign.
\end{summarybox}

\subsection{Quantitative Results and Ablation Study}\label{quant_comp}

\begin{figure*}
    \centering
    \includegraphics[width=0.9\textwidth]{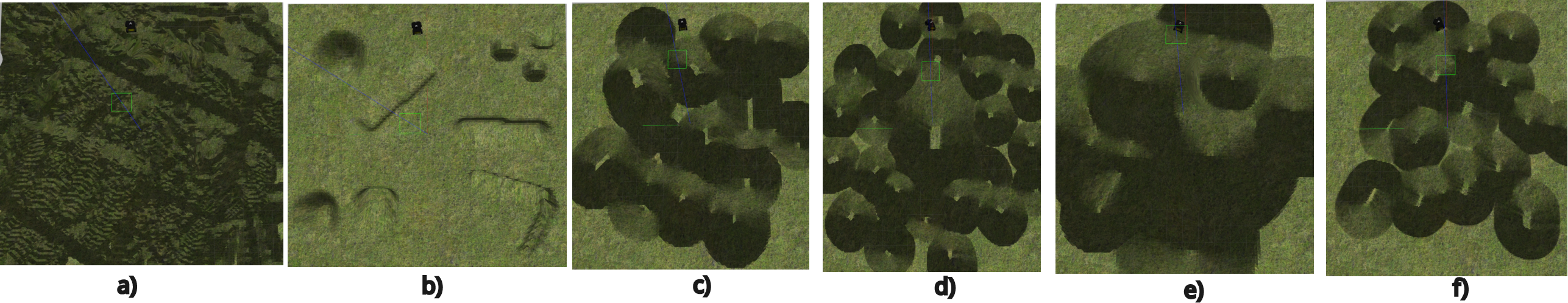}
    \caption{Synthetic terrains used for quantitative benchmarking. a) Terrain 1. b) Terrain 2. c) Terrain 3. d) Terrain 4. e) Terrain 5. f) Terrain 6.}
    \label{fig:terrain_maps}
\end{figure*}

\noindent We evaluated closed-loop navigation performance on the six synthetic terrains shown in Fig.~\ref{fig:terrain_maps}. Four planner configurations were tested on the same 30 randomized start--goal pairs for each terrain: the complete proposed framework, Bi-level-opt, GP-navigation, and an ablation with $c_{unc}=0$. This yields 180 trials per configuration and 720 trials in total, of which 540 form the three-method baseline comparison and 180 form the complete-versus-ablation comparison. A run is classified as a failure if the vehicle does not reach the goal within the allotted time, becomes immobilized, or exceeds a roll or pitch magnitude of $40^\circ$.

Table~\ref{tab:simulation_stats} summarizes the results. The complete proposed framework produced 34 failures in 180 trials, corresponding to an aggregate failure rate of $18.9\%$. Bi-level-opt produced 83 failures ($46.1\%$), while GP-navigation produced 75 failures ($41.7\%$). The complete framework, therefore, reduced the observed failure rate by 27.2 percentage points relative to Bi-level-opt and by 22.8 percentage points relative to GP-navigation. It attained the lowest or tied-lowest failure rate on every terrain and the strictly lowest rate on five of the six terrains; the only tie occurred with Bi-level-opt on Terrain~1.

The fourth row for each terrain reports the ablation, which sets $c_{unc}=0$ in \eqref{eqn:final_cem_cost}. It retains the same terrain representation, NLS pose/contact predictor, nominal objective, CEM optimizer, start--goal pairs, and evaluation settings as the complete framework. The ablation produced 62 failures in 180 trials, giving a failure rate of $34.4\%$, which is 15.6 percentage points above the complete proposed framework; it also produced more failures on each of the six terrains.

The aggregate results reflect the different ways in which the planners account for missing terrain observations. Bi-level-opt evaluates the NLS-predicted pose and wheel contacts on a deterministic terrain fit, allowing sparsely supported extrapolations to receive favorable nominal costs. GP-navigation excludes cells whose predictive variance exceeds a prescribed threshold. The proposed framework instead assigns a graded uncertainty penalty that depends on the predicted body pose, individual wheel contacts, and surface normals at those contacts. This vehicle-configuration-dependent treatment is consistent with the lower failure rate achieved by the proposed planner across the matched benchmark.

The attitude statistics in Table~\ref{tab:simulation_stats} are computed over successful trials and therefore characterize traversal quality after completion. These values vary across terrains, and no planner dominates every roll and pitch metric. We consequently use failure rate as the primary measure of reliability, with RMS and peak attitude providing complementary information about successful traversals.

\begin{summarybox}
Across the 180 matched trials, the complete framework achieved the lowest aggregate failure rate at $18.9\%$, a reduction of 27.2 percentage points relative to Bi-level-opt and 22.8 percentage points relative to GP-navigation. It also produced fewer failures than the ablation on every terrain, reducing the aggregate failure rate from $34.4\%$ to $18.9\%$. Together with the controlled comparison in Section~\ref{uncert_effect}, these results show that propagated-uncertainty scoring changes route selection under matched CEM initialization, while the complete pipeline provides the strongest aggregate closed-loop performance. Also, see Appendix \ref{sec:appendix_paired_ablation} for additional ablation, which quantifies the effect of incorporating uncertainty.
\end{summarybox}

\begin{table*}
\centering
\caption{Simulation Statistics Across Terrains. A run is classified as a failure if the vehicle does not reach the goal within the allotted time, becomes immobilized, or exceeds a roll or pitch magnitude of $40^\circ$. The planner formulation represents pose angles in radians internally; roll and pitch are converted to degrees here for ease of interpretation. The RMS and Max values are taken from successful runs only. ``Ablation'' sets $c_{unc}=0$.}
\label{tab:simulation_stats}
\small
\renewcommand{\arraystretch}{1.2}
\begin{tabular}{ll ccc cccc}
\toprule
\multirow{2}{*}{\textbf{Terrain}} & \multirow{2}{*}{\textbf{Method}} & \multirow{2}{*}{\textbf{Runs}} & \multirow{2}{*}{\textbf{Failures}} & \multirow{2}{*}{\textbf{Failure Rate (\%)}} & \multicolumn{2}{c}{\textbf{Roll ($^\circ$)}} & \multicolumn{2}{c}{\textbf{Pitch ($^\circ$)}} \\
\cmidrule(lr){6-7} \cmidrule(lr){8-9}
& & & & & \textbf{RMS} & \textbf{Max} & \textbf{RMS} & \textbf{Max} \\
\midrule
\multirow{4}{*}{\textbf{Terrain 1}} & \textbf{Proposed} & 30 & 1 & \textbf{3.33} & 3.962 & 28.068 & 5.419 & 29.159 \\
 & Bi-level-opt & 30 & 1 & \textbf{3.33} & 6.440 & 31.468 & 6.313 & 34.315 \\
 & GP-navigation & 30 & 2 & 6.67 & 2.523 & 21.332 & 6.954 & 38.693 \\
 & \shortstack{Ablation ($c_{unc}=0$)} & 30 & 2 & 6.67 & 4.905 & 28.588 & 5.991 & 35.009 \\
 \midrule
\multirow{4}{*}{\textbf{Terrain 2}} & \textbf{Proposed} & 30 & 12 & \textbf{40.00} & 4.708 & 31.927 & 2.581 & 33.775 \\
 & Bi-level-opt & 30 & 25 & 83.33 & 8.702 & 39.285 & 6.838 & 32.306 \\
 & GP-navigation & 30 & 25 & 83.33 & 5.398 & 22.806 & 6.662 & 33.345 \\
 & \shortstack{Ablation ($c_{unc}=0$)} & 30 & 23 & 76.67 & 3.553 & 35.974 & 4.986 & 38.694 \\
 \midrule
\multirow{4}{*}{\textbf{Terrain 3}} & \textbf{Proposed} & 30 & 4 & \textbf{13.33} & 10.096 & 37.150 & 10.471 & 34.312 \\
 & Bi-level-opt & 30 & 17 & 56.67 & 11.502 & 38.472 & 12.189 & 39.713 \\
 & GP-navigation & 30 & 12 & 40.00 & 7.793 & 37.762 & 9.089 & 35.229 \\
 & \shortstack{Ablation ($c_{unc}=0$)} & 30 & 7 & 23.33 & 10.250 & 30.657 & 12.545 & 36.010 \\
 \midrule
\multirow{4}{*}{\textbf{Terrain 4}} & \textbf{Proposed} & 30 & 15 & \textbf{50.00} & 11.383 & 38.561 & 12.656 & 39.675 \\
 & Bi-level-opt & 30 & 23 & 76.67 & 14.910 & 36.608 & 21.221 & 39.012 \\
 & GP-navigation & 30 & 19 & 63.33 & 9.312 & 39.142 & 11.588 & 38.359 \\
 & \shortstack{Ablation ($c_{unc}=0$)} & 30 & 19 & 63.33 & 15.907 & 39.767 & 12.698 & 36.928 \\
 \midrule
\multirow{4}{*}{\textbf{Terrain 5}} & \textbf{Proposed} & 30 & 2 & \textbf{6.67} & 7.786 & 36.227 & 12.046 & 38.408 \\
 & Bi-level-opt & 30 & 13 & 43.33 & 11.346 & 39.131 & 13.559 & 34.713 \\
 & GP-navigation & 30 & 10 & 33.33 & 7.302 & 27.753 & 7.795 & 34.190 \\
 & \shortstack{Ablation ($c_{unc}=0$)} & 30 & 7 & 23.33 & 10.379 & 39.332 & 11.928 & 38.846 \\
 \midrule
\multirow{4}{*}{\textbf{Terrain 6}} & \textbf{Proposed} & 30 & 0 & \textbf{0.00} & 4.753 & 24.437 & 6.481 & 24.374 \\
 & Bi-level-opt & 30 & 4 & 13.33 & 8.266 & 26.054 & 7.723 & 28.507 \\
 & GP-navigation & 30 & 7 & 23.33 & 4.306 & 18.824 & 7.144 & 34.894 \\
 & \shortstack{Ablation ($c_{unc}=0$)} & 30 & 4 & 13.33 & 8.491 & 25.958 & 7.796 & 27.078 \\
\bottomrule
\end{tabular}
\end{table*}

\subsection{Hardware Results}\label{real_world_demo}
\begin{figure}
    \centering
    \includegraphics[width=\columnwidth]{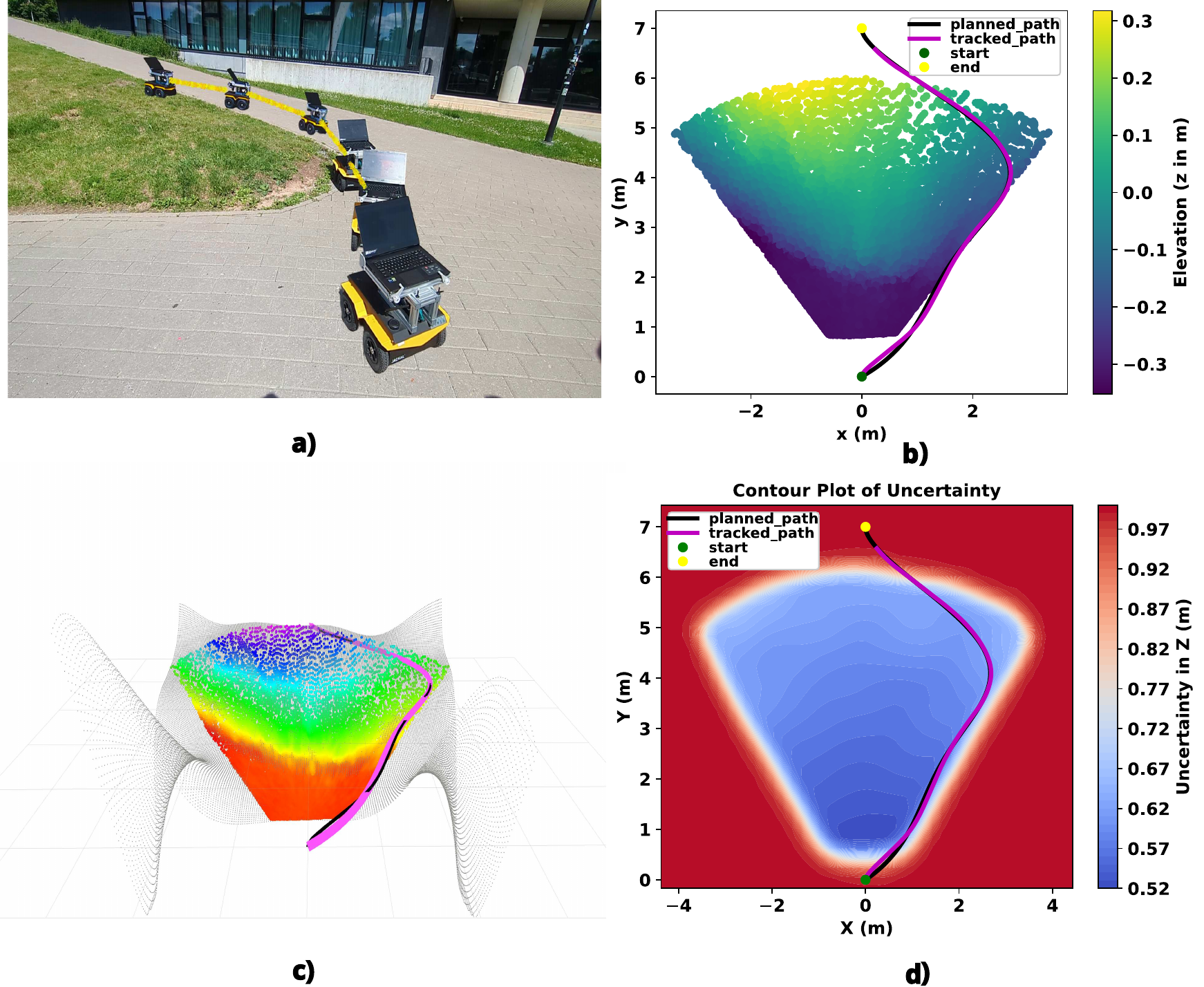}
    \caption{Illustrative Clearpath Jackal execution with varying observation support and modeled terrain traversability. (a)~Outdoor scene and robot execution. (b)~Planned trajectory (black) and tracked trajectory (magenta) overlaid on the partial point cloud. (c)~Fitted terrain and point-cloud observations; the direct start--goal route crosses the elevated grassy region. (d)~Coverage-induced terrain-uncertainty map. The direct route is observed but has a higher modeled surface-normal cost; the selected route moves toward the boundary of the observed footprint, where this cost is lower, before turning toward the goal.}
    \label{fig:outdoor5_figs}
\end{figure}

\noindent Hardware evaluations were conducted in six distinct outdoor environments. We present two representative executions here; executions from the remaining four environments are included in the supplementary video. Figure~\ref{fig:tree} reports a Clearpath Husky execution, whereas Fig.~\ref{fig:outdoor5_figs} reports a Clearpath Jackal execution. In both figures, the planned trajectory is shown in black, and the trajectory tracked by the robot is shown in magenta. The first shows route selection near a coverage-induced blind spot; the second, route selection where observation support and modeled terrain traversability both vary.

Figure~\ref{fig:tree} shows the Clearpath Husky navigating past a tree that occludes part of the terrain from the depth camera. The dashed cyan ellipse in panels~(b) and~(d) marks the weakly observed region behind the tree. The selected route remains within the better-observed corridor and avoids the marked occluded region, behavior consistent with the intended coverage-uncertainty penalty. Panels~(a) and~(c) show the physical scene and fitted terrain, respectively, while panel~(b) qualitatively compares the planned and tracked trajectories.
\begin{figure}
    \centering
    \includegraphics[width=\columnwidth]{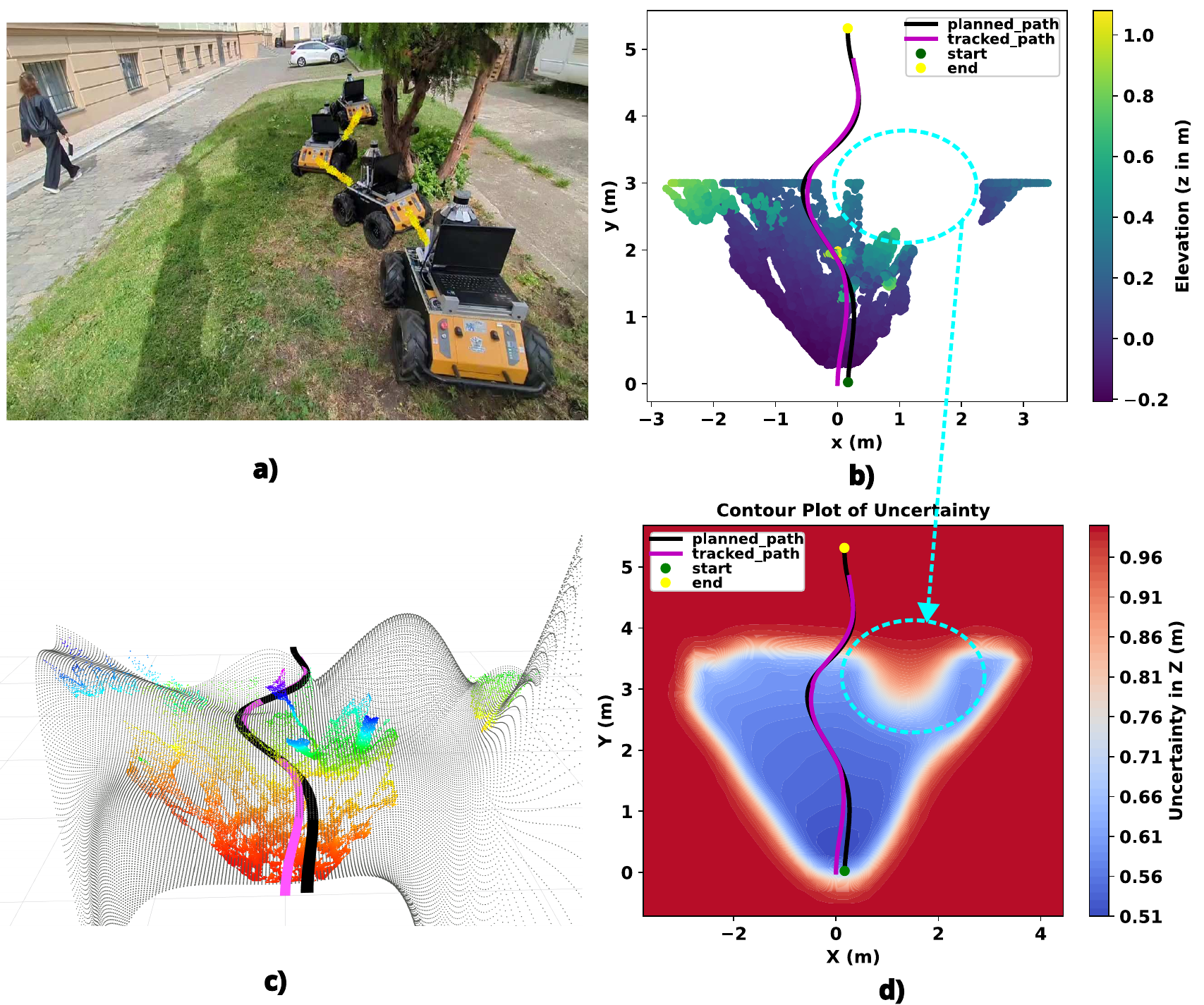}
    \caption{Illustrative Clearpath Husky execution near a coverage-induced blind spot. (a)~Outdoor scene and robot execution. (b)~Planned trajectory (black) and tracked trajectory (magenta) overlaid on the partial point cloud. The dashed cyan ellipse marks the region occluded by the tree. (c)~Fitted terrain and point-cloud observations. (d)~Coverage-induced terrain-uncertainty map. The selected route remains in the better-observed corridor and routes around the occlusion.}
    \label{fig:tree}
\end{figure}

Figure~\ref{fig:outdoor5_figs} presents the complementary case. A straight start--goal path would remain inside the observed region, as seen in panel~(d), and would therefore appear favorable if coverage uncertainty were considered alone. However, the fitted terrain in panel~(c) shows that this route would cross the elevated grassy region and produce an unfavorable surface-normal cost in \eqref{eqn:normal_cost}. The selected trajectory moves toward the boundary of the observed footprint, where the modeled surface-normal cost is lower, before returning toward the goal. This behavior is consistent with the combined objective. Panel~(b) qualitatively compares the planned and tracked trajectories.

\begin{summarybox}
Together, Figs.~\ref{fig:tree} and~\ref{fig:outdoor5_figs} demonstrate end-to-end hardware feasibility and show route choices consistent with the intended roles of observation support and modeled vehicle--terrain interaction costs. 
\end{summarybox}

\subsection{Performance of the Terrain Parameter Network} \label{terrain_fm}
\noindent We evaluated the Terrain Parameter Network as an amortized warm start for local terrain fitting. The model was trained on 2,866 spatially disjoint local point-cloud patches and evaluated on a held-out set of 319 spatially disjoint patches, each corresponding to the sensor-limited terrain window used by the planner. For every test patch, one Gaussian source sample was integrated through the Flow Matching ODE to obtain a single Fourier-parameter prediction, matching the deployment procedure. Fitting RMSE was measured between the fitted surface elevations and the reference elevations in the corresponding point-cloud patch. We evaluated the direct network prediction, the same prediction followed by exactly one Levenberg--Marquardt refinement iteration, and the nonlinear solver run to convergence.

Table~\ref{tab:terrain_param_prediction} reports the mean per-patch fitting RMSE and computation time over all 319 test patches. All methods were evaluated on an NVIDIA RTX 5090 GPU, and timing was performed after JIT compilation. Direct network prediction required 3\,ms, compared with 43\,ms for the converged nonlinear solver, yielding a $14.3\times$ speedup with an absolute fitting RMSE increase of 0.06\,mm ($3.1\%$). One refinement iteration recovered the nonlinear solver's fitting RMSE at the reported precision while requiring 14\,ms, retaining a $3.1\times$ speedup.

\begin{table}[!t]
    \centering
    \caption{Terrain-parameter prediction performance averaged over 319 spatially disjoint held-out local point-cloud patches. One Flow Matching sample is used per patch, and computation time is measured on an NVIDIA RTX 5090 after JIT compilation. Lower values are better.}
    \label{tab:terrain_param_prediction}
    \renewcommand{\arraystretch}{1.15}
    \begin{tabular}{@{}lcc@{}}
        \toprule
        \textbf{Method} & \textbf{Fitting RMSE (mm)} & \textbf{Time (ms)} \\
        \midrule
        Network Prediction & 1.99 & 3 \\
        One-Step Refinement & 1.93 & 14 \\
        Nonlinear Solver & 1.93 & 43 \\
        \bottomrule
    \end{tabular}
\end{table}

\begin{summarybox}
These results show that the network provides an accurate initialization for the model-based terrain fit: direct prediction offers the lowest latency, while one refinement iteration matches the solver's reported fitting RMSE at substantially lower computational cost.
\end{summarybox}

%% file: Conclusions.tex
\section{Conclusions}

This paper presented an occlusion-aware framework for generating quasi-static, stability-oriented reference trajectories for rigid, non-articulated four-wheeled vehicles on uneven terrain from partial point-cloud observations. A regularized inverse-Hessian estimate in a fixed-feature Fourier terrain model represents the coverage-induced epistemic uncertainty created by sensor blind spots. Implicit differentiation and first-order propagation transfer this estimate through the NLS wheel--terrain interaction model to the predicted pose, wheel contacts, and per-wheel surface-normal costs. CEM combines these uncertainty penalties with the nominal geometric objective, assigning additional cost to trajectories whose predicted vehicle--terrain interaction is sensitive to weakly supported terrain.

Across six synthetic terrains, the complete framework attained an aggregate observed failure rate of $18.9\%$, compared with $46.1\%$ for Bi-level-opt and $41.7\%$ for GP-navigation, corresponding to reductions of 27.2 and 22.8 percentage points. The $c_{\mathrm{unc}}=0$ ablation had a failure rate of $34.4\%$, 15.6 percentage points above the complete framework. Hardware evaluations across six distinct outdoor environments provided qualitative evidence of end-to-end planning and closed-loop execution on physical wheeled robots; two representative executions are presented in the paper, and four additional executions are included in the supplementary video.

The Flow Matching model provides a warm start without replacing model-based evaluation. Direct terrain-parameter prediction was $14.3\times$ faster than the converged nonlinear solver with a $3.1\%$ increase in fitting RMSE; one Levenberg--Marquardt refinement iteration recovered the solver's reported fitting RMSE while remaining $3.1\times$ faster. Thus, learning shifts computation offline without replacing the analytical terrain fit or the downstream CEM planning.

The framework assumes a rigid, non-articulated four-wheeled platform, a single-valued terrain height field, quasi-static vehicle--terrain interaction, and local first-order uncertainty propagation. Geometric hazards are represented through the terrain model and soft traversability costs rather than through separate hard constraints. Establishing formal guarantees under partial observation on highly uneven terrain requires calibrated uncertainty models and constrained reasoning through nonlinear vehicle--terrain interactions, which can be computationally demanding for online planning. The present work, therefore, adopts a tractable formulation that minimizes nominal geometric costs together with propagated uncertainty penalties. Stronger probabilistic guarantees could be pursued through a chance-constrained formulation, albeit with additional modeling requirements and computational overhead. Future work will also incorporate dynamic stability and slip, quantify performance over larger repeated hardware campaigns, and use active viewpoint selection to reduce blind-spot uncertainty before the robot commits to a partially observed passage.

\appendix[Implementation Details]\label{sec:appendix_parameters}

\subsection{Planner Weights and Numerical Stabilizers}

Table~\ref{tab:planner_weights_stabilizers} collects the cost weights, uncertainty-penalty coefficients, and numerical stabilizers used by the planner.

\begin{table}[!h]
\centering
\caption{Planner weights, penalty coefficients, and numerical stabilizers.}
\label{tab:planner_weights_stabilizers}
\footnotesize
\renewcommand{\arraystretch}{1.12}
\begin{tabular}{@{}l >{\raggedright\arraybackslash}p{0.43\columnwidth} c l@{}}
\toprule
\textbf{Symbol} & \textbf{Role} & \textbf{Value} & \textbf{Unit} \\
\midrule
$w_{\kappa}$ & Curvature-cost weight & 0.01 & $\mathrm{m}^{2}$ \\
$w_a$ & Acceleration-cost weight & 10.0 & $\mathrm{s}^{4}\mathrm{m}^{-2}$ \\
$w_n$ & Surface-normal-cost weight & 10.0 & dimensionless \\
$w_p$ & Pose-cost weight & 10.0 & $\mathrm{rad}^{-2}$ \\
$\rho_n$ & Normal-uncertainty penalty coefficient & 1.0 & dimensionless \\
$\rho_p$ & Pose-uncertainty penalty coefficient & 1.0 & dimensionless \\
$\epsilon$ & Curvature denominator stabilizer & $10^{-6}$ & $\mathrm{m}^{2}\mathrm{s}^{-2}$ \\
$\epsilon_v$ & Low-speed yaw threshold & 0.01 & $\mathrm{m}\,\mathrm{s}^{-1}$ \\
\bottomrule
\end{tabular}
\end{table}

Table~\ref{tab:planner_evaluation_protocol} reports the remaining vehicle, solver, randomization, and closed-loop evaluation settings used in the experiments.

\begin{table}[!h]
\centering
\caption{Planner and closed-loop evaluation protocol.}
\label{tab:planner_evaluation_protocol}
\footnotesize
\renewcommand{\arraystretch}{1.08}
\begin{tabular}{@{}>{\raggedright\arraybackslash}p{0.25\columnwidth}
                    >{\raggedright\arraybackslash}p{0.40\columnwidth}
                    >{\raggedright\arraybackslash}p{0.25\columnwidth}@{}}
\toprule
\textbf{Category} & \textbf{Required setting} & \textbf{Value} \\
\midrule
Vehicle & Wheelbase $l$, width $w$, and wheel offsets $h_i$ & $l{=}0.210$\,m, $w{=}0.272$\,m, $h_i{=}0.260$\,m ($i{=}1,\ldots,4$) \\
CEM sampling & Initial coefficient mean and covariance & boundary-consistent straight-line Bernstein coefficients; empirical covariance of Bernstein fits to 100 straight-line paths with smoothness-shaped lateral noise, $+\,10^{-3}\,\mathbf{I}$ \\
Terrain LM & Damping, stopping tolerance, and iteration limit & damping $10^{-6}$; gradient tol.\ $10^{-3}$; 1 iteration (online refinement) \\
Pose/contact NLS & Stopping tolerance and iteration limit & gradient-norm tol.\ $10^{-3}$; 30 iterations \\
Scenario generation & Start--goal sampling law and admissibility rules & goal uniform in $90^{\circ}$ forward cone, 6--8\,m range, inside map \\
Goal condition & Goal radius and closed-loop episode timeout & 0.75\,m; 120\,s \\
Immobilization & Speed/displacement threshold and evaluation window & displacement $<0.20$\,m over 20\,s window, checked at 20\,Hz \\
Execution & Controller-update rate & 20\,Hz \\
Sensor geometry & Sensor pose and horizontal/vertical field of view & arch-mounted LiDAR: $360^{\circ}/30^{\circ}$; front depth camera: $87^{\circ}/58^{\circ}$ \\
Environment & Collision and map-boundary handling & Gazebo contact physics; goals confined to the map interior \\
Outcome labeling & Precedence when multiple failure events occur & timeout, then immobilization, then goal (20\,Hz); post-hoc $40^{\circ}$ attitude check \\
\bottomrule
\end{tabular}
\end{table}

\subsection{Flow Matching Network Details}\label{sec:appendix_fm_details}

This subsection documents the data provenance, numerical training configuration, and inference settings of the Flow Matching network. Table~\ref{tab:fm_reproducibility_details} reports the settings used for training and evaluation.

\begin{table}[!h]
\centering
\caption{Flow Matching training and inference details.}
\label{tab:fm_reproducibility_details}
\footnotesize
\renewcommand{\arraystretch}{1.08}
\begin{tabular}{@{}>{\raggedright\arraybackslash}p{0.20\columnwidth}
                    >{\raggedright\arraybackslash}p{0.24\columnwidth}
                    >{\raggedright\arraybackslash}p{0.45\columnwidth}@{}}
\toprule
\textbf{Category} & \textbf{Required information} & \textbf{Value} \\
\midrule
Terrain data & Point clouds and split & 3{,}185 patches of 35{,}000 points; shuffled 90/10 split (2{,}866/319) \\
Terrain network & CNN/DiT dimensions and size & CNN 64--128--256 $\to$ 256; DiT width 256, depth 4, 8 heads, MLP 1024; 5.47\,M parameters \\
Normalization & Input/target scaling & each input shifted/scaled by one training-set mean and std.; targets (Fourier coefficients) in raw units \\
Optimization & Optimizer, lr, batch, epochs, precision & AdamW, weight decay $10^{-2}$, constant lr: $10^{-3}$, batch 1{,}000, $\le$10{,}000 epochs, float32 \\
Model selection & Checkpoint criterion & lowest training loss \\
ODE inference & Integrator and batching & fixed-step explicit midpoint; 16 uniform steps on $[0,1]$; batched by \texttt{vmap} \\
Randomization & Seeds & seed 0 for initialization and loader shuffling; fixed per-instance PRNG keys for FM source noise \\
\bottomrule
\end{tabular}
\end{table}

\subsection{Paired Outcomes for the Uncertainty Ablation}
\label{sec:appendix_paired_ablation}

Because the complete framework and the ablation were evaluated on the same 180 terrain/start--goal scenarios, their marginal failure rates can be decomposed into the paired outcomes in Table~\ref{tab:paired_uncertainty_ablation}. Each scenario is classified by the outcome of the complete framework, followed by that of the ablation, preserving the within-scenario comparison that is hidden by aggregate failure counts.

\begin{table}[!h]
\centering
\caption{Paired outcomes for the complete framework and the $c_{\mathrm{unc}}=0$ ablation. The complete framework outcome is listed first; S and F denote success and failure, respectively.}
\label{tab:paired_uncertainty_ablation}
\footnotesize
\renewcommand{\arraystretch}{1.12}
\begin{tabular}{@{}lcccc@{}}
\toprule
\textbf{Outcome (complete/ablation)} & \textbf{S/S} & \textbf{S/F} & \textbf{F/S} & \textbf{F/F} \\
\midrule
\textbf{Scenarios} & 108 & 38 & 10 & 24 \\
\bottomrule
\end{tabular}
\end{table}

The two configurations agreed on 132 scenarios: both succeeded in 108, and both failed in 24. Their outcomes differed on the remaining 48 scenarios. The complete framework succeeded while the ablation failed in 38 cases, whereas the reverse occurred in 10 cases. The 28-scenario difference between these discordant outcomes exactly accounts for the reduction from 62 failures for the ablation to 34 for the complete framework, corresponding to the 15.6-percentage-point gap reported in Table~\ref{tab:simulation_stats}. Thus, under matched terrain and start--goal conditions, outcome reversals favored the complete framework substantially more often than the ablation.

%% file: references.bib
@inproceedings{peebles2023scalable,
  title={Scalable diffusion models with transformers},
  author={Peebles, William and Xie, Saining},
  booktitle={Proceedings of the IEEE/CVF international conference on computer vision},
  pages={4195--4205},
  year={2023}
}

@article{tong2023conditional,
  title={Conditional flow matching: Simulation-free dynamic optimal transport},
  author={Tong, Alexander and Malkin, Nikolay and Huguet, Guillaume and Zhang, Yanlei and Rector-Brooks, Jarrid and Fatras, Kilian and Wolf, Guy and Bengio, Yoshua},
  journal={arXiv preprint arXiv:2302.00482},
  volume={2},
  number={3},
  year={2023}
}

@article{lipman2024flow,
  title={Flow matching guide and code},
  author={Lipman, Yaron and Havasi, Marton and Holderrieth, Peter and Shaul, Neta and Le, Matt and Karrer, Brian and Chen, Ricky TQ and Lopez-Paz, David and Ben-Hamu, Heli and Gat, Itai},
  journal={arXiv preprint arXiv:2412.06264},
  year={2024}
}

@inproceedings{carvalho2023motion,
  title={Motion Planning Diffusion: Learning and Planning of Robot Motions with Diffusion Models},
  author={Carvalho, Jo{\~a}o and Le, An T. and Baierl, Mark and Koert, Dorothea and Peters, Jan},
  booktitle={2023 IEEE/RSJ International Conference on Intelligent Robots and Systems (IROS)},
  pages={1916--1923},
  year={2023},
  doi={10.1109/IROS55552.2023.10342382},
  organization={IEEE}
}

@inproceedings{nguyen2025flowmp,
  title={FlowMP: Learning Motion Fields for Robot Planning with Conditional Flow Matching},
  author={Nguyen, Khang and Le, An T. and Pham, Tien and Huber, Manfred and Peters, Jan and Vu, Minh Nhat},
  booktitle={2025 IEEE/RSJ International Conference on Intelligent Robots and Systems (IROS)},
  pages={11291--11297},
  year={2025},
  doi={10.1109/IROS60139.2025.11246537},
  organization={IEEE}
}

@article{chakraborty2004kinematics,
  title={Kinematics of wheeled mobile robots on uneven terrain},
  author={Chakraborty, Nilanjan and Ghosal, Ashitava},
  journal={Mechanism and machine theory},
  volume={39},
  number={12},
  pages={1273--1287},
  year={2004},
  publisher={Elsevier}
}

@article{singh2016feasible,
  title={Feasible acceleration count: A novel dynamic stability metric and its use in incremental motion planning on uneven terrain},
  author={Singh, Arun Kumar and Krishna, K Madhava},
  journal={Robotics and Autonomous Systems},
  volume={79},
  pages={156--171},
  year={2016},
  publisher={Elsevier}
}

@article{papadopoulos2000force,
  title={The force-angle measure of tipover stability margin for mobile manipulators},
  author={Papadopoulos, Evangelos and Rey, Daniel A},
  journal={Vehicle System Dynamics},
  volume={33},
  number={1},
  pages={29--48},
  year={2000},
  publisher={Taylor \& Francis}
}

@inproceedings{manoharan2024bi,
  title={Bi-level trajectory optimization on uneven terrains with differentiable wheel-terrain interaction model},
  author={Manoharan, Amith and Sharma, Aditya and Belsare, Himani and Pal, Kaustab and Krishna, K Madhava and Singh, Arun Kumar},
  booktitle={International Conference on Intelligent Robots and Systems (IROS)},
  pages={11977--11984},
  year={2024},
  organization={IEEE}
}

@misc{blender,
  title = {{Blender}},
  howpublished = "\url{https://www.blender.org/}"
}

@misc{jax2018github,
  author = {James Bradbury and Roy Frostig and Peter Hawkins and Matthew James Johnson and Yash Katariya and Chris Leary and Dougal Maclaurin and George Necula and Adam Paszke and Jake Vander{P}las and Skye Wanderman-{M}ilne and Qiao Zhang},
  title = {{JAX}: composable transformations of {P}ython+{N}um{P}y programs},
  url = {http://github.com/jax-ml/jax},
  version = {0.3.13},
  year = {2018},
}

@inproceedings{koenig2004design,
  title={Design and use paradigms for gazebo, an open-source multi-robot simulator},
  author={Koenig, Nathan and Howard, Andrew},
  booktitle={IEEE/RSJ international conference on intelligent robots and systems},
  pages={2149--2154},
  year={2004}
}

@article{jaxopt_implicit_diff,
  title={Efficient and Modular Implicit Differentiation},
  author={Blondel, Mathieu and Berthet, Quentin and Cuturi, Marco and Frostig, Roy 
    and Hoyer, Stephan and Llinares-L{\'o}pez, Felipe and Pedregosa, Fabian 
    and Vert, Jean-Philippe},
  journal={arXiv preprint arXiv:2105.15183},
  year={2021}
}

@inproceedings{quigley2009ros,
  title={ROS: an open-source Robot Operating System},
  author={Quigley, Morgan and Conley, Ken and Gerkey, Brian and Faust, Josh and Foote, Tully and Leibs, Jeremy and Wheeler, Rob and Ng, Andrew Y.},
  booktitle={ICRA workshop on open source software},
  year={2009}
}

@article{kidger2021equinox,
    author={Patrick Kidger and Cristian Garcia},
    title={{E}quinox: neural networks in {JAX} via callable {P}y{T}rees and filtered transformations},
    year={2021},
    journal={Differentiable Programming workshop at Neural Information Processing Systems}
}

@misc{jia2025m2udmultimodelmultiscenariouneventerrain,
      title={M2UD: A Multi-model, Multi-scenario, Uneven-terrain Dataset for Ground Robot with Localization and Mapping Evaluation}, 
      author={Yanpeng Jia and Shiyi Wang and Shiliang Shao and Yue Wang and Fu Zhang and Ting Wang},
      year={2025},
      eprint={2503.12387},
      archivePrefix={arXiv},
      primaryClass={cs.RO},
      url={https://arxiv.org/abs/2503.12387}, 
}

@inproceedings{leininger2024gaussian,
  title={Gaussian process-based traversability analysis for terrain mapless navigation},
  author={Leininger, Abe and Ali, Mahmoud and Jardali, Hassan and Liu, Lantao},
  booktitle={International Conference on Robotics and Automation (ICRA)},
  pages={10925--10931},
  year={2024},
  organization={IEEE}
}

@ARTICLE{pi-mppi,
  author={Andrejev, Edvin Martin and Manoharan, Amith and Unt, Karl-Eerik and Singh, Arun Kumar},
  journal={IEEE Robotics and Automation Letters}, 
  title={$\pi$-MPPI: A Projection-Based Model Predictive Path Integral Scheme for Smooth Optimal Control of Fixed-Wing Aerial Vehicles}, 
  year={2025},
  volume={10},
  number={6},
  pages={6496-6503},
  doi={10.1109/LRA.2025.3568323}}

@misc{clearpathrobotics,
  title = {{Clearpath Robotics}},
  howpublished = "\url{https://clearpathrobotics.com/}"
}

@misc{realsense_d435,
  title = {{Intel RealSense Depth Camera D435}},
  howpublished = "\url{https://www.realsenseai.com/products/stereo-depth-camera-d435/}"
}

@inproceedings{todorov2012mujoco,
  title={MuJoCo: A physics engine for model-based control},
  author={Todorov, Emanuel and Erez, Tom and Tassa, Yuval},
  booktitle={International Conference on Intelligent Robots and Systems},
  pages={5026--5033},
  year={2012},
  organization={IEEE},
  doi={10.1109/IROS.2012.6386109}
}

@misc{agx,
  title = {{
AGX Dynamics
}},
  howpublished = "\url{https://www.algoryx.se/agx-dynamics/}"
}

@article{Wang_2025,
doi = {10.1088/1361-6501/add28d},
url = {https://doi.org/10.1088/1361-6501/add28d},
year = {2025},
month = {may},
publisher = {IOP Publishing},
volume = {36},
number = {6},
pages = {066303},
author = {Wang, Han and Wang, Gongcheng and Zhu, Hongbiao and Du, Chengjin and Bai, Hua and Ding, Pengchao and Xu, Wenda and Du, Zhijiang and Wang, Weidong},
title = {A robust navigation framework for uneven terrain with traversability-focused planning},
journal = {Measurement Science and Technology}
}

@article{hu2026safe,
  title={Safe and Robust Terrain Vehicle Navigation Based on Risk-Aware Path-Planning and Control},
  author={Hu, Chuan and Wang, Zhidong and Wang, Ziao and Niu, Yixun and Taghavifar, Hamid and Yin, Jianhua and Qin, Yechen},
  journal={IEEE Transactions on Intelligent Vehicles},
  year={2026}
}

@article{kim2024pts,
  title={PTS-Map: Probabilistic Terrain State Map for Uncertainty-Aware Traversability Mapping in Unstructured Environments},
  author={Kim, Dong-Wook and Son, E-In and Kim, Chan and Hwang, Ji-Hoon and Seo, Seung-Woo},
  journal={IEEE Robotics and Automation Letters},
  volume={10},
  number={2},
  pages={1257--1264},
  year={2024}
}

@article{park2025cute,
  title={CUTE-Planner: Confidence-aware Uneven Terrain Exploration Planner},
  author={Park, Miryeong and Cho, Dongjin and Kim, Sanghyun and Cho, Younggun},
  journal={arXiv preprint arXiv:2511.12984},
  year={2025}
}

@inproceedings{tazaki2025slope,
  title={Slope-aware Maximum Uncertainty Sampling for Time-efficient Robotic Mapping on Rough Terrain},
  author={Tazaki, Minori and Ishigami, Genya},
  booktitle={International Conference on Space Robotics (iSpaRo)},
  pages={64--71},
  year={2025},
  organization={IEEE}
}

@article{pairet2021online,
  title={Online mapping and motion planning under uncertainty for safe navigation in unknown environments},
  author={Pairet, {\`E}ric and Hern{\'a}ndez, Juan David and Carreras, Marc and Petillot, Yvan and Lahijanian, Morteza},
  journal={IEEE Transactions on Automation Science and Engineering},
  volume={19},
  number={4},
  pages={3356--3378},
  year={2021}
}

@inproceedings{tan2025real,
  title={Real-time spatial-temporal traversability assessment via feature-based sparse gaussian process},
  author={Tan, Senming and Hou, Zhenyu and Zhang, Zhihao and Xu, Long and Zhang, Mengke and He, Zhaoqi and Xu, Chao and Gao, Fei and Cao, Yanjun},
  booktitle={IEEE/RSJ International Conference on Intelligent Robots and Systems (IROS)},
  pages={17533--17540},
  year={2025}
}

@inproceedings{lee2023learning,
  title={Learning-based uncertainty-aware navigation in 3d off-road terrains},
  author={Lee, Hojin and Kwon, Junsung and Kwon, Cheolhyeon},
  booktitle={International Conference on Robotics and Automation (ICRA)},
  pages={10061--10068},
  year={2023},
  organization={IEEE}
}

@inproceedings{triest2024unrealnet,
  title={Unrealnet: Learning uncertainty-aware navigation features from high-fidelity scans of real environments},
  author={Triest, Samuel and Fan, David D and Scherer, Sebastian and Agha-Mohammadi, Ali-Akbar},
  booktitle={IEEE International Conference on Robotics and Automation (ICRA)},
  pages={12627--12634},
  year={2024}
}

@article{endo2026deep,
  title={Deep probabilistic traversability with test-time adaptation for uncertainty-aware planetary rover navigation},
  author={Endo, Masafumi and Taniai, Tatsunori and Ishigami, Genya},
  journal={Scientific reports},
  volume={16},
  number={1},
  pages={9499},
  year={2026},
  publisher={Nature Publishing Group UK London}
}

@article{rubinstein1999cross,
  title={The cross-entropy method for combinatorial and continuous optimization},
  author={Rubinstein, Reuven},
  journal={Methodology and computing in applied probability},
  volume={1},
  number={2},
  pages={127--190},
  year={1999},
  publisher={Springer}
}

@inproceedings{xu2023efficient,
  title={An efficient trajectory planner for car-like robots on uneven terrain},
  author={Xu, Long and Chai, Kaixin and Han, Zhichao and Liu, Hong and Xu, Chao and Cao, Yanjun and Gao, Fei},
  booktitle={2023 IEEE/RSJ International Conference on Intelligent Robots and Systems (IROS)},
  pages={2853--2860},
  year={2023},
  organization={IEEE}
}

@article{benatti2022end,
  title={End-to-end learning for off-road terrain navigation using the Chrono open-source simulation platform},
  author={Benatti, Simone and Young, Aaron and Elmquist, Asher and Taves, Jay and Tasora, Alessandro and Serban, Radu and Negrut, Dan},
  journal={Multibody System Dynamics},
  volume={54},
  number={4},
  pages={399--414},
  year={2022},
  publisher={Springer}
}

@article{yin2023reliable,
  title={Reliable global path planning of off-road autonomous ground vehicles under uncertain terrain conditions},
  author={Yin, Jianhua and Li, Lingxi and Mourelatos, Zissimos P and Liu, Yixuan and Gorsich, David and Singh, Amandeep and Tau, Seth and Hu, Zhen},
  journal={IEEE Transactions on Intelligent Vehicles},
  volume={9},
  number={1},
  pages={1161--1174},
  year={2023},
  publisher={IEEE}
}

@article{wiberg2021control,
  title={Control of rough terrain vehicles using deep reinforcement learning},
  author={Wiberg, Viktor and Wallin, Erik and Nordfjell, Tomas and Servin, Martin},
  journal={IEEE robotics and automation letters},
  volume={7},
  number={1},
  pages={390--397},
  year={2021},
  publisher={IEEE}
}

@article{fan2021step,
  title={STEP: Stochastic Traversability Evaluation and Planning for Risk-Aware Off-road Navigation},
  author={Fan, David D and Otsu, Kyohei and Kubo, Yuki and Dixit, Anushri and Burdick, Joel and Agha-Mohammadi, Ali-Akbar},
  journal={Robotics: Science and Systems XVII},
  year={2021},
  publisher={Robotics: Science and Systems Foundation}
}

@inproceedings{beyer2024risk,
  title={Risk-predictive planning for off-road autonomy},
  author={Beyer, Lukas Lao and Ryou, Gilhyun and Spieler, Patrick and Karaman, Sertac},
  booktitle={2024 IEEE International Conference on Robotics and Automation (ICRA)},
  pages={16452--16458},
  year={2024},
  organization={IEEE}
}

@inproceedings{castro2023does,
  title={How does it feel? self-supervised costmap learning for off-road vehicle traversability},
  author={Castro, Mateo Guaman and Triest, Samuel and Wang, Wenshan and Gregory, Jason M and Sanchez, Felix and Rogers, John G and Scherer, Sebastian},
  booktitle={IEEE International Conference on Robotics and Automation (ICRA)},
  pages={931--938},
  year={2023},
  organization={IEEE}
}

@article{bhosale2026review,
  title={A Review of Learning Off-Road Terrain Traversability for Autonomous Ground Vehicles},
  author={Bhosale, Mayuresh and Whitson, Jordan A and Jia, Yunyi},
  journal={Journal of Autonomous Vehicles and Systems},
  pages={1--24},
  year={2026}
}

@article{frey2024roadrunner,
  title={RoadRunner--Learning traversability estimation for autonomous off-road driving},
  author={Frey, Jonas and Patel, Manthan and Atha, Deegan and Nubert, Julian and Fan, David and Agha, Ali and Padgett, Curtis and Spieler, Patrick and Hutter, Marco and Khattak, Shehryar},
  journal={IEEE Transactions on Field Robotics},
  volume={1},
  pages={192--212},
  year={2024},
  publisher={IEEE}
}

@article{waibel2022rough,
  title={How rough is the path? Terrain traversability estimation for local and global path planning},
  author={Waibel, Gabriel G{\"u}nter and L{\"o}w, Tobias and Nass, Mathieu and Howard, David and Bandyopadhyay, Tirthankar and Borges, Paulo Vinicius Koerich},
  journal={IEEE Transactions on Intelligent Transportation Systems},
  volume={23},
  number={9},
  pages={16462--16473},
  year={2022},
  publisher={IEEE}
}

@article{jun2016pose,
  title={Pose estimation-based path planning for a tracked mobile robot traversing uneven terrains},
  author={Jun, Jae-Yun and Saut, Jean-Philippe and Benamar, Faiz},
  journal={Robotics and Autonomous Systems},
  volume={75},
  pages={325--339},
  year={2016},
  publisher={Elsevier}
}

@article{yu2026real,
  title={Real-Time Topology-Aware Local Planning and Control for Off-Road Vehicles on 3-D Terrains},
  author={Yu, Siyuan and Shen, Congkai and Epureanu, Bogdan I and Ersal, Tulga},
  journal={IEEE Transactions on Control Systems Technology},
  year={2026},
  publisher={IEEE}
}

@inproceedings{fabian2020pose,
  title={Pose prediction for mobile ground robots in uneven terrain based on difference of heightmaps},
  author={Fabian, Stefan and Kohlbrecher, Stefan and Von Stryk, Oskar},
  booktitle={2020 IEEE International Symposium on Safety, Security, and Rescue Robotics (SSRR)},
  pages={49--56},
  year={2020},
  organization={IEEE}
}

@inproceedings{jian2022putn,
  title={Putn: A plane-fitting based uneven terrain navigation framework},
  author={Jian, Zhuozhu and Lu, Zihong and Zhou, Xiao and Lan, Bin and Xiao, Anxing and Wang, Xueqian and Liang, Bin},
  booktitle={2022 IEEE/RSJ International Conference on Intelligent Robots and Systems (IROS)},
  pages={7160--7166},
  year={2022},
  organization={IEEE}
}

@inproceedings{tian2023efficient,
  title={An efficient planning framework for ground vehicles navigating on uneven terrain},
  author={Tian, Xiaohui and Yang, Shuaicong and Yu, Kai and Fu, Mengyin and Song, Wenjie},
  booktitle={2023 IEEE International Conference on Unmanned Systems (ICUS)},
  pages={1493--1498},
  year={2023},
  organization={IEEE}
}

@article{hua2024double,
  title={Double neural networks enhanced global mobility prediction model for unmanned ground vehicles in off-road environments},
  author={Hua, Chen and Jiang, Chunmao and Niu, Runxin and Fu, Xuhui and Chen, Ziyu and Kuang, Xinkai and Yu, Biao},
  journal={IEEE Transactions on Vehicular Technology},
  volume={73},
  number={6},
  pages={7547--7560},
  year={2024},
  publisher={IEEE}
}

@inproceedings{jordan2017real,
  title={Real-time pose estimation on elevation maps for wheeled vehicles},
  author={Jordan, Julian and Zell, Andreas},
  booktitle={2017 IEEE/RSJ International Conference on Intelligent Robots and Systems (IROS)},
  pages={1337--1342},
  year={2017},
  organization={IEEE}
}

@article{ning2026ut,
  title={UT-Planner: Energy-Efficient Trajectory Planner for Unmanned Ground Vehicles on Uneven Terrain},
  author={Ning, Changjiu and Sun, Chao and Yang, Xiongji and Huang, Zhishuai and Wen, Da and Chen, Zitong and Leng, Jianghao},
  journal={IEEE Internet of Things Journal},
  year={2026},
  publisher={IEEE}
}

@article{datar2024learning,
  title={Learning to model and plan for wheeled mobility on vertically challenging terrain},
  author={Datar, Aniket and Pan, Chenhui and Xiao, Xuesu},
  journal={IEEE Robotics and Automation Letters},
  volume={10},
  number={2},
  pages={1505--1512},
  year={2024},
  publisher={IEEE}
}

@inproceedings{weerakoon2022terp,
  title={Terp: Reliable planning in uneven outdoor environments using deep reinforcement learning},
  author={Weerakoon, Kasun and Sathyamoorthy, Adarsh Jagan and Patel, Utsav and Manocha, Dinesh},
  booktitle={International Conference on Robotics and Automation (ICRA)},
  pages={9447--9453},
  year={2022},
  organization={IEEE}
}
